\documentclass[letterpaper, 10 pt, conference]{ieeeconf}  % Comment this line out
\IEEEoverridecommandlockouts                              % This command is only
\usepackage[pdftex]{graphicx}
\usepackage[cmex10]{amsmath}
\usepackage[utf8]{inputenc}
\usepackage{amssymb}
\usepackage{import}
\usepackage{bm}

\usepackage{color}
\usepackage{units}
\usepackage{url}
\usepackage{float}
\usepackage{epstopdf}
\usepackage{stmaryrd}

\usepackage{units}
\usepackage{siunitx}

\usepackage{outlines}
\usepackage{hyperref}
\usepackage{lipsum}

\usepackage{boldline}
\usepackage[flushleft]{threeparttable}
\usepackage{todonotes}
\usepackage{color,soul}

\usepackage{mathtools}
\usepackage{subcaption}
\usepackage{multirow}
\usepackage{multicol}
\usepackage{algorithm}% http://ctan.org/pkg/algorithms
\usepackage{algorithmicx}
\usepackage{algpseudocode}

\usepackage{epstopdf}

\usepackage{etoolbox}\AtBeginEnvironment{algorithmic}{\small}

\newcommand{\mnoshow}[1]{}

\newcommand{\figref}[1]{Fig.~\ref{#1}}
\newcommand{\secref}[1]{Section~\ref{#1}}

\newtheorem{remark}{Remark}

\newcommand{\norm}[1]{\left\lVert#1\right\rVert}

\newcommand{\boldlambda}{\boldsymbol{\lambda}}
\newcommand{\boldmu}{\boldsymbol{\mu}}

\title{\LARGE \bf
Formation and Reconfiguration of Tight Multi-Lane Platoons}

\author{Roya Firoozi, Xiaojing Zhang, Francesco Borrelli
\thanks{The authors are with the Department of Mechanical Engineering, at the University of California, Berkeley.{\tt \{royafiroozi, xiaojing.zhang, fborrelli \}@berkeley.edu}}%
}

\begin{document}
\maketitle
\thispagestyle{empty}
\pagestyle{empty}
%%%%%%%%%%%%%%%%%%%%%%%%%%%%%%%%%%%%%%%%%%%%%%%%%%%%%%%%%%%%%%%%%%%%%%%%%%%%%%%%
\begin{abstract}
Advances in vehicular communication technologies are expected to facilitate cooperative driving in the future. Connected and Automated Vehicles (CAVs) are able to collaboratively plan and execute driving maneuvers by sharing their perceptual knowledge and future plans. In this paper, an architecture for autonomous navigation of tight multi-lane platoons travelling on public roads is presented. Using the proposed approach, CAVs are able to form single or multi-lane platoons of various geometrical configurations. They are able to reshape and adjust their configurations according to changes in the environment. The proposed architecture consists of two main components: an offline motion planner system and an online hierarchical control system.
The motion planner uses an optimization-based approach for cooperative formation and reconfiguration in tight spaces. A constrained optimization scheme is used to plan smooth, dynamically feasible and collision-free trajectories for all the vehicles within the platoon. The paper addresses online computation limitations by employing a family of maneuvers precomputed offline and stored on a look-up table on the vehicles. The online hierarchical control system is composed of three levels: a traffic operation system (TOS), a decision-maker, and a path-follower. The TOS determines the desired platoon reconfiguration. The decision-maker checks the feasibility of the reconfiguration plan based on real-time information about the surrounding traffic. The reconfiguration maneuver is executed by a low-level path-following feedback controller in real-time. The effectiveness of the approach is demonstrated through simulations of three case studies: 1) formation reconfiguration 2) obstacle avoidance, and 3) benchmarking against behavior-based planning in which the desired formation is achieved using a sequence of motion primitives. Videos and software can be found online here \tt{\url{https://github.com/RoyaFiroozi/Centralized-Planning}}. 
\end{abstract}

\section{Introduction}
Vehicular wireless communication systems including vehicle-to-vehicle (V2V), vehicle-to-cloud (V2C) and vehicle to infrastructure (V2I) enhance cooperative driving by providing a communication network for information exchange between the vehicles to coordinate and plan conflict-free trajectories \cite{GUANETTI201818}, \cite{Alam2010AnES}. Grouping multiple cooperative vehicles into single-lane or multi-lane formation is referred to as platooning. Using communication technologies, connected vehicles within the platoon can navigate in close proximity of each other, self-organize themselves to form certain configurations, keep tight formations and transit from one formation to another. Platooning improves traffic congestion, energy efficiency and safety \cite{Alam2015}, \cite{SUN201937}. It increases road traffic throughput by allowing small inter-vehicle distances. Furthermore, moving with close spacing reduces aerodynamic drag and thus contributes to energy efficiency.

Platooning in classical setting refers to a group of vehicles that form a road train in a single lane \cite{Hedrick1991}, \cite{Shladover1991}. Single-lane platooning study and demonstrations date back to the '80s \cite{Rajamani2000DesignAE}, \cite{HanShue1998}. The main drawback of forming a single-lane platoon is that a long train-like platoon may prevent other vehicles to change lane and consequently affect the traffic flow and reduce the mobility. Also in case of presence of obstacles on the road it might be impossible for a long platoon to find enough gap to change lane. Platoon formation in multiple lanes incorporates the advantages of platooning described earlier and at the same time is shape-reconfigurable and is able to facilitate lane change maneuvers as needed. In this paper, a multi-lane platoon with small number of interconnected vehicles (three up to ten) referred to as \emph{mini-platoon} is considered. Adding another degree of freedom in multi-lane platoon increases structure flexibility and can further improve mobility, the traffic network throughput, energy efficiency and safety compared to single-lane platoon. For example, in terms of energy efficiency, when there is slow traffic ahead in one lane, multi-lane platoon can reconfigure its shape and perform opportunistic lane change to save the energy consumption by avoiding braking and changing the lane to a faster lane \cite{Gueriau}. In terms of safety, once an obstacle is detected in one lane, the multi-lane platoon can reconfigure and accommodate the vehicles in the blocked lane to merge into another lane to avoid the obstacle and minimize the risk of possible collision.

Although single-lane platooning (one-dimensional 1D) is well studied in the literature, literature on multi-lane platoons (two-dimensional 2D) is limited and reviewed in the next section. The focus of this paper is to present a general architecture for autonomous navigation of tight multi-lane platoons. The contributions are summarized as follows.
\begin{enumerate}
    \item An \emph{architecture} for autonomous navigation of multi-lane platoons on public roads is presented. It comprises an \emph{offline motion-planning system} and an \emph{online hierarchical control system}. 
    \item A set of formation patterns also referred to as single-lane and multi-lane platoon configurations is identified. The offline motion planner uses an optimization-based algorithm to create various reconfiguration maneuvers that allow smooth transitioning from one pre-identified configuration to another. The resulting reconfiguration maneuvers are stored in a look-up table.
    
    \item The online hierarchical control system is composed of three levels: a \emph{traffic operation system (TOS)}, a \emph{decision-maker}, and a \emph{path-follower}. The top level TOS operates in the cloud and determines the desired platoon reconfiguration by monitoring the traffic. The middle-level decision-maker operates on the platoon leader vehicle. It makes use of the following information:
    \begin{itemize}
        \item the desired reconfiguration from TOS, via V2C communication,
        \item the look-up table computed by motion-planner, pre-stored on the vehicles, 
        \item and the shared future plans of the surrounding traffic (outside platoon) vehicles, via V2V communication.
    \end{itemize}
    By incorporating all these information, the decision-maker checks whether the desired reconfiguration planned by TOS is feasible or not. The feasible maneuvers are broadcasted to all the vehicles within the platoon via V2V communication to be executed by the low-level path-following feedback controller in real-time.
    \item The vehicles' shapes are modeled as polytopic sets and the collision avoidance constraints among them are reformulated into a set of smooth constraints using strong duality theory. These smooth constraints can be handled efficiently by standard non-linear solvers. This approach allows navigation through tight spaces at highway speed.
    \item Compared to existing literature, the three novel contributions discussed above address real-time implementation, tight maneuvering and hard constraint satisfaction. Uncertainty is not addressed in this work and is topic of ongoing research.
\end{enumerate}

The remainder of the paper is structured as follows. \secref{secLiteratureReview} provides a literature review about multi-vehicle formation. \secref{secPreliminaries} describes preliminaries. \secref{secMotionPlanning} presents the proposed motion planning approach, and describes the decision-making and planning scheme structure. \secref{secSimpleExample} introduces motion planning using sequence of motion primitives, which is used as a benchmark to compare the proposed planning approach against. \secref{secResults} presents simulation results and \secref{secConclusion} concludes the paper and presents future research directions.

\section{Literature Review}
\label{secLiteratureReview}

Coordinated formation methods for multiple autonomous vehicles are well-studied in the literature and can be categorized in three main approaches: Leader-follower, virtual structure, and behavior-based approach. In leader-follower approach the follower agents track the coordinates of the leader \cite{7124428}, \cite{inbook}. This method is effective for conventional single-lane train-like platoon, but since the follower must follow the same reference trajectory as the leader, it is not applicable to reconfigurable multi-lane platoons, in which the planned motions for the vehicles are not the same. 
In virtual structure method, the formation is represented as a virtual rigid structure. Each robot is considered as a node in the rigid structure \cite{inproceedings}. The main drawback of this method is that, the formation as a rigid structure is not flexible and reshapable.     

Behavior-based approaches include methodologies such as flocking and particle swarm optimization algorithms, artificial potential fields, and sequence of motion primitives. Most of the studies on flocking algorithms consider the agents as a group of particles that interact with each other based on Reynolds heuristic rules of cohesion, separation and alignment \cite{Reynolds:1987:FHS:37402.37406}. Cohesion enforces the particles to stay together and separation penalizes the collision between the particles. 
In artificial potential field method, potential fields are built so that the robot is attracted by the goal region and repelled by the obstacle region. In formation control, in addition to goal and obstacle potential fields, a swarm attractive field is introduced to achieve the desired formation pattern. The potential-based planning does not impose hard constraint on collision avoidance and cannot guarantee collision avoidance with constrained control input. In addition, all these particle-based methods model the vehicles as particles with radial gap among them and do not take the actual size of the vehicles into account. Furthermore, the dynamic model is considered to be the particle's dynamic with first, second or third-order point-mass models, which are not the representation of the actual nonlinear dynamics of the vehicles.

Another behavior-based method is to construct the formation maneuvers as sequences of motion primitives \cite{736776}. Motion primitives are identified as various behaviors such as lane change and obstacle avoidance. Among all the described formation approaches, this method is more effective for multi-lane platooning, but its disadvantage is that it is difficult to mathematically analyze and solve for sequence of motion primitives.

Combinations of the aforementioned approaches have also been studied.
In \cite{1605401}, for example, the authors use the Reynolds rules to define the potential forces between the agents. Cohesion and separation are modeled as pairwise attractive and repulsive potential forces between the particle, respectively and a multi-objective cost function is constructed to satisfy all the rules simultaneously. In \cite{980728}, the authors propose virtual leader approach with attractive potential field to track a desired path and achieve a desired formation and repulsive potential fields to avoid agents collisions. Also a Lyapunov function is constructed to prove the closed-loop stability. In \cite{7749193}, the authors use a similar approach for flocking of multiple non-holonomic vehicles and prove the convergence using LaSalle's invariant principle.

\section{Preliminaries}
\label{secPreliminaries}
\subsection{Vehicle Model}
The vehicles set composing  the platoon is defined as $\mathcal V$. The number of vehicles are considered to be $N_v$ and each vehicle is identified through its index $i\in\mathcal V := \{1,2,...,N_v\}$.
The nonlinear behavior of every vehicle $i$ within the set is modeled by the vehicle kinematic bicycle model, which is a common modeling approach in path planning. 
\begin{figure}[h]
\centering
\includegraphics[scale=0.4]{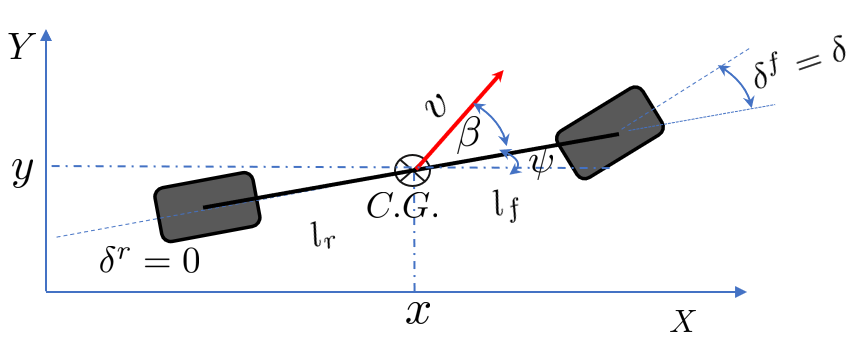}
\caption{The kinematic bicycle model }
\label{fig:kinematic_bicycle_model}
\end{figure}
In this model, the $i$th vehicle state vector is $\mathbf{z}^i = [x^i,y^i,\psi^i,v^i]^\top$, where $x^i$ and $y^i$ represent longitudinal and lateral positions of the vehicle, respectively, $\psi^i$ is the heading angle and $v^i$ denotes the velocity at center of gravity (C.G.) of the vehicle, as seen in \figref{fig:kinematic_bicycle_model}. The control input vector is defined as $\mathbf{u}^i = [a^i, \delta^i]^\top$, where $a^i$ is the acceleration and $\delta^i$ is the steering angle. The vehicle dynamics is given as follows 
\begin{equation}\label{eq:kinematic_bicycle_model}
\begin{aligned}
\dot x^i & = v^i \cos(\psi^i+\beta^i),  \\
\dot y^i & = v^i \sin(\psi^i+\beta^i), \\
\dot \psi^i &= \frac{v^i\cos\beta^i}{l_f^i+l_r^i}(\tan\delta^i),\\
\dot v^i &= a^i,
\end{aligned}
\end{equation}
where $\beta^i = \arctan\big(\tan\delta^i(\frac{l_r^i}{l_f^i+l_r^i})\big)$ is the side slip angle, $l_{f}^i$ and $l_{r}^i$ are the distance from the center of gravity to the front and rear axles, respectively. Superscript $i$ in this paper denotes the $i$th vehicle in the platoon. Using Euler discretization, the model \eqref{eq:kinematic_bicycle_model} is discretized as follows
\begin{equation}\label{eq:kinematic_bicycle_model_discrete}
\begin{aligned}
x^i(t+1) & = x^i(t) + \Delta t \ v^i(t) \cos(\psi^i(t)+\beta^i(t)),  \\
y^i(t+1) & = y^i(t)+\Delta t \ v^i(t)  \sin(\psi^i(t)+\beta^i(t)), \\
\psi^i(t+1) &= \psi^i(t)+ \Delta t \ \frac{v^i(t)\cos\beta^i(t)}{l_f^i+l_r^i}(\tan\delta^i(t)),\\
v^i(t+1) &= v^i(t) + \Delta t \ a^i(t),
\end{aligned}
\end{equation}
where $\Delta t$ is the  sampling time.
\subsection{Platoon Configuration}
\begin{figure}
\centering
\includegraphics[scale=0.5]{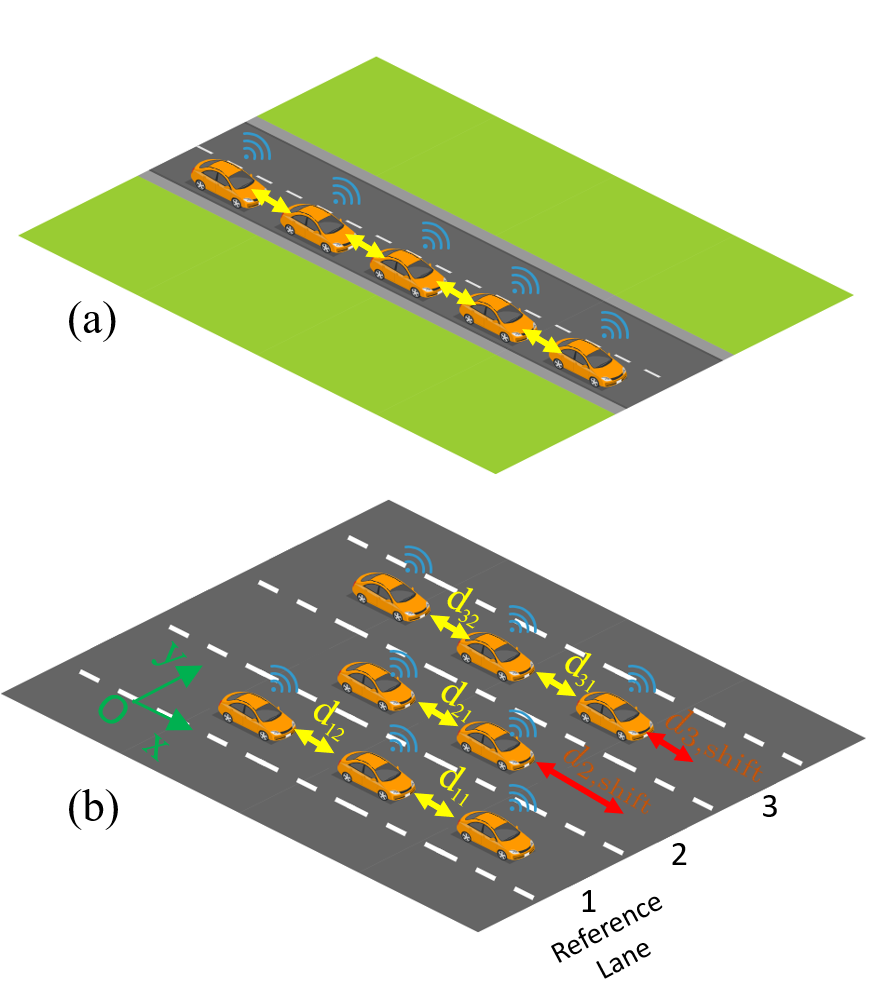}
\caption{(a) Single-lane platoon: a train-like group of vehicles travelling at close distance behind each other. (b) Multi-lane platoon in multiple lanes. Yellow arrow depicts horizontal inter-vehicle distance at each lane and red arrow shows $d_{j,\text{shift}}$ at each lane. Each lane has its own label and the right-most lane is the reference lane.}
\label{fig:platoon_configurations}
\end{figure}

Various platoon formation patterns or configurations are considered in this work, including one-lane (train-like) and multi-lane (rectangle, diamond, wedge shape, etc.), as shown in \figref{fig:platoon_configurations}. The platoon configuration $\mathcal{C}$ is parameterized as $\mathcal{C}(n_{v},l,p)$, where $n_v \in \mathbb{Z}$ is the maximum number of vehicles in each lane within the platoon, $l \in \{0,1\}^{n_l}$ is an indicator vector that specifies which lanes are occupied, $n_l$ is the maximum number of lanes within the platoon. The $j$th element of $l$ is defined as 
\begin{align*}
l(j) = 
\begin{cases}
      0 & \text{if no vehicle is in $j$th lane} \\
      1 &  \text{if at least one vehicle is in $j$th lane,}
\end{cases} 
\end{align*}
where $j$ denotes the lane index. The parameter matrix $p \in \mathbb{R}^{n_l \times n_v}$ represents the platoon geometrical pattern specified as the relative distances between the vehicles. Every $j$th row of matrix $p$ is defined as $p(j) = [d_{j,\text{shift}}, d_{j1},..., d_{j(n_v-1)}]$, where $d_{j1},..., d_{j(n_v-1)}$ denote the horizontal inter-vehicle distances at $j$th lane as shown in \figref{fig:platoon_configurations}(b) and $d_{j,\text{shift}}$ is the horizontal shifting distance of the front-most vehicle at each lane with respect to the front end of the reference vehicle. The right-most lane in direction of travel is the reference lane for $j$th lane, as shown in \figref{fig:platoon_configurations}(b) and the reference vehicle is the front-most vehicle at reference lane. For the cars ahead of the reference vehicle, $d_{j,\text{shift}}$ is considered as negative. The values of $d_{j1},..., d_{j(n_v-1)}$ and $d_{j,\text{shift}}$ are design parameters and might be chosen as different values for each lane. For example, the platoon configuration in \figref{fig:platoon_configurations}(b) is defined as 
\begin{align*}
\mathcal{C} = \mathcal{C}(3,[1,1,1],p), \
p = \begin{bmatrix}
0& 1& 1\\
2& 1& 0\\
1& 1& 1
\end{bmatrix},
\end{align*}
where $d_{1,\text{shift}}$, $d_{2,\text{shift}}$ and $d_{3,\text{shift}}$ associated with 1st, 2nd and 3rd lanes are $0m$, $2m$ and $1m$, respectively. Also $d_{11}, d_{12}, d_{21}, d_{31}$ and $d_{32}$ are all $1m$ in this configuration.  
For trajectory optimization purposes, it is convenient to convert the configuration $\mathcal{C}$ to position coordinates $(x,y)^i$ of each vehicle $i$ within the platoon. The function
% \begin{align}\label{eq:g}
% g: & \mathbb N \times \mathbb \{0,1\}^{n_l} \times \mathbb R^{2\times n_l}\rightarrow \mathbb R^{2N_v},\\ 
% & \mathcal{C}(n_{v},l,p)\rightarrow (z^1,\ldots, z^{N_v})
% \end{align}
\begin{equation}\label{eq:g}
g:\mathcal{C}(n_{v},l,p)\rightarrow ((x,y)^1,\ldots, (x,y)^{N_v}),
\end{equation}
gets the configuration $\mathcal{C}$ as input and outputs the position coordinates $(x,y)$ for all the vehicles. The origin $O$, as shown in \figref{fig:platoon_configurations}, is defined as the position of the rear-most vehicle at the right-most lane of the platoon configuration and all the coordinates are determined with respect to that origin. 

\subsection{Simple Reference Generator  Model}
A simple integrator function is defined which is used in \secref{secMotionPlanning}, to generate the reference trajectories for each vehicle. The function $h: \mathbb R^3 \to \mathbb R^T,$ is defined as
% $(x(0),v_{\text{max}},T)\mapsto x_{\text{Ref}}$ 
\begin{equation}\label{eq:h}
h: (x(0),v_{\text{max}},T) \to % x_{\text{Ref}}^{i}
% h(x^i(0),v_{\text{max}},T) = 
x_{\text{Ref}} = [x(0), x(1),...,x(T)],
\end{equation} 
which determines $x_{\text{Ref}}$ for all the vehicles within the platoon. The trajectory is obtained by $x(t+1) = x(t) + v_{\text{max}}\Delta t,\quad \forall{t} \in \{0,1,...,T\}$, where $x(t)$ is the vehicle longitudinal position at time $t$, $v_{\text{max}}$ is the maximum speed limit of the road, $T$ is the final time of simulation and $\Delta t$ is the simulation sampling time. 

\subsection{Platoon Reconfiguration}
\begin{figure}[ht]
\centering
\includegraphics[scale = 0.5]{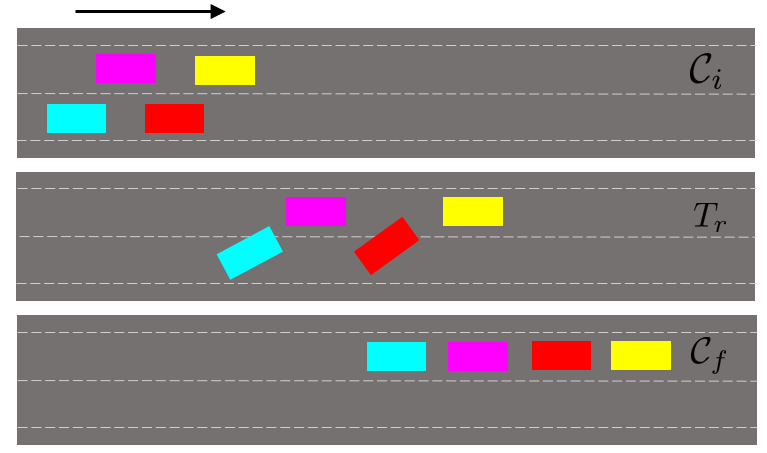}
\caption{An example of reconfiguration from multi-lane platoon to single-lane platoon is shown.}
\label{fig:reconf_example}
\end{figure}

Transitioning from an initial configuration denoted as $\mathcal{C}_i$ to a final configuration denoted as $\mathcal{C}_f$ is defined as platoon reconfiguration. An example of platoon reconfiguration is shown in \figref{fig:reconf_example}. The top snapshot shows a multi-lane platoon with initial configuration $\mathcal{C}_i$ that is going forward at steady state (right-headed arrow shows the direction of motion). The middle snapshot shows transition maneuvers $T_r$ and the vehicles change their lanes. Whenever the transition maneuver is completed, another configuration $\mathcal{C}_f$, which in this example is single-lane platoon, is achieved as shown in the bottom snapshot. A finite number of platoon configurations are identified as known configurations. The configuration set $\mathbb{C}$ = $\{\mathcal{C}_1, \mathcal{C}_2, \hdots\}$ captures all these pre-defined platoon configurations. The platoon reconfiguration scenarios are restricted to transition between these pre-defined configurations.

\begin{remark}
The traffic operation system (TOS) selects the desired platoon configurations among all the pre-identified configurations within the set $\mathbb{C}$, in such a way to improve traffic mobility and to reduce traffic congestion. The vehicles communicate with this level via V2C communication and receive $\mathcal{C}_i$ and $\mathcal{C}_f$. The platoon might not always be initially in a pre-identified $\mathcal{C}_i$ configuration, due to the changes in the surrounding traffic. Therefore, to initiate the reconfiguration, the vehicles are first controlled to reach $\mathcal{C}_i$ configuration. After reaching such a pre-defined configuration, the reconfiguration maneuver is initiated. Single-lane (1D) platoon formation from an unknown configuration has been studied for a long time. One way to reach the $\mathcal{C}_i$ is to first form a simple 1D platoon and initiate the reconfiguration from that known simple platoon. Another way (recommended in this paper) is to control the vehicles to reach $\mathcal{C}_i$. Each vehicle individually plans and controls to reach its corresponding location and when reaching the goal is not possible, due to the surrounding traffic condition, the vehicle informs TOS that reaching to $\mathcal{C}_i$ is infeasible and then TOS re-plans the reconfiguration.
\end{remark}

\subsection{Surrounding Traffic}
The vehicles set composing the surrounding traffic (the vehicles travelling close to the platoon, but do not belong to the platoon) is defined as the set $\mathcal{S}=\{1,...,n\}$, where $n$ is the total number of the surrounding vehicles. Each vehicle is identified through its index $q \in \mathcal{S}$. These vehicles are in the communication range of the platoon and share their future planned trajectories with the platoon leader. (The front-most vehicle in the reference lane within the platoon, is chosen as platoon leader.)

\begin{figure}[ht]
\centering
\includegraphics[scale = 0.25]{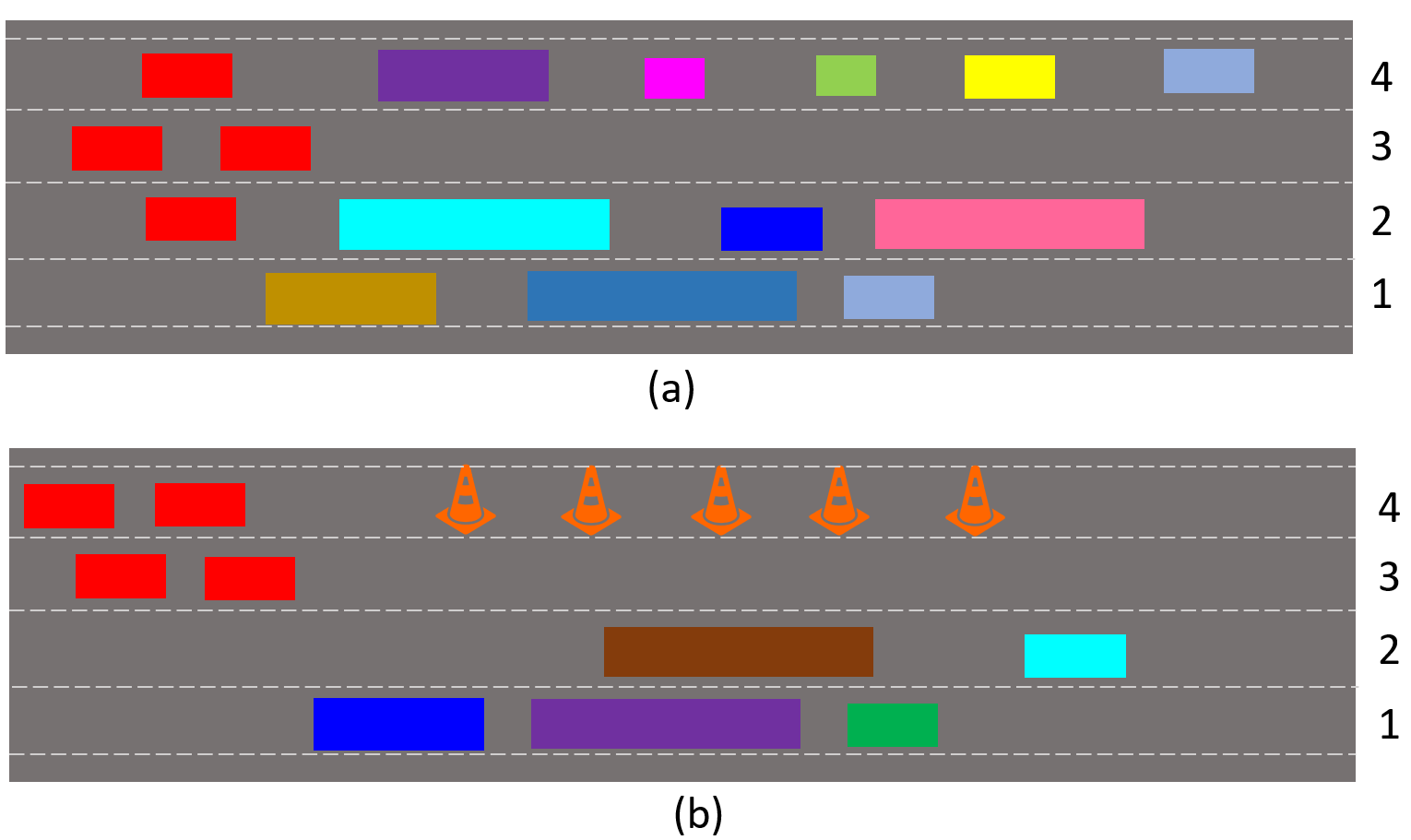}
\caption{The vehicles within the platoon are shown in red and the surrounding traffic (vehicles not in platoon) are shown with different colors. (a): The traffic is slow in the lanes 2 and 4. (b): The lane 4 is closed.}
\label{fig:traffic_scenarios}
\end{figure}

Different traffic scenario examples are shown in \figref{fig:traffic_scenarios}. In these examples, the multi-lane platoon (shown in red) can reconfigure to improve the traffic flow. In \figref{fig:traffic_scenarios}(a), a three-lane platoon is moving in the lanes 2,3,4. Since the traffic is slow in the lanes 2 and 4, a possible reconfiguration is that platoon can merge into the lane 3 and reconfigure as a single-lane platoon. In \figref{fig:traffic_scenarios}(b), the lane 4 is closed due to an accident, the platoon vehicles in the lane 4 can merge between the platoon vehicles in the lane 3.

\subsection{Notations}
Common used notations along with their definitions are reported in the Table \ref{tab:notations}. The configurations are denoted using $\mathcal{C}$ and the trajectories are denoted using $\tau$. The superscription $i$ indicates $ith$ vehicle.

\begin{table}
\centering
\begin{tabular}{|c|c|}
\hline
{\textbf{Notation}}& {\textbf{Definition}} \\
\hline
$\mathcal{C}$ & platoon configuration\\
$\mathcal{C}_i$ & initial platoon configuration\\
$\mathcal{C}_f$ & final platoon configuration\\
$\mathbb{C}$ & the set of pre-defined/known configurations\\
$N_v$ & number of vehicles\\
$\mathcal{S}$ & surrounding traffic vehicles (not in platoon) \\
$\mathcal{V}$ & set of all the vehicles\\
$i$ & index of $ith$ vehicle\\
$\mathcal{N}_i$ & set of neighbor vehicles (within platoon) of $ith$ vehicle\\
$\mathbf{z}^i$ & states of $ith$ vehicle\\
$\mathbf{u}^i$ & inputs of $ith$ vehicle\\
$x^i$ & longitudinal position of $ith$ vehicle\\
$y^i$ & lateral position of $ith$ vehicle\\
$\psi^i$ & heading angle of $ith$ vehicle\\
$v^i$ & velocity of $ith$ vehicle\\
$\mathbf{z}^i_{\text{Ref}}$ & reference states of $ith$ vehicle\\
$\mathcal{P}$ & polytopic representation of the vehicle\\
$T$ & final simulation (maneuver) time\\
$\mathbf{R}$ & rotation matrix\\
$\mathbf{t}_r$ & translation vector\\
$\mathbf{A}$ and $\mathbf{b}$ & polytopic representation\\ 
$len$ & the vehicle length\\
$w$ & the vehicle width\\
$d_{\text{min}}$ & minimum safe distance\\
$t$ & time step\\
$k$ & horizon step\\
$N$ & horizon\\
$\boldlambda$, $\boldmu$, $\mathbf{s}$ & dual variables\\
$\boldsymbol{\tau}$ & trajectory\\
$\boldsymbol{\tau}^i_{\mathbf{z}_{\textrm{Ref}}}$ & $ith$ vehicle reference state trajectory\\
$\boldsymbol{\tau}^i_{\mathbf{z}_{\textrm{Target}}}$ & $ith$ vehicle target trajectory (look-up table)\\
$\boldsymbol{\tau}^q_{\mathbf{z}}$ & shared planned trajectory of $qth$ surrounding vehicle \\
$\rho$ & coefficient affecting the start of lane-change, $\rho \in (0,1)$\\
\hline
\end{tabular}
\caption{Common used notations}
\label{tab:notations}
\end{table}

\section{Architecture}
\label{secMotionPlanning}
\begin{figure}[t]
\includegraphics[width = \linewidth]{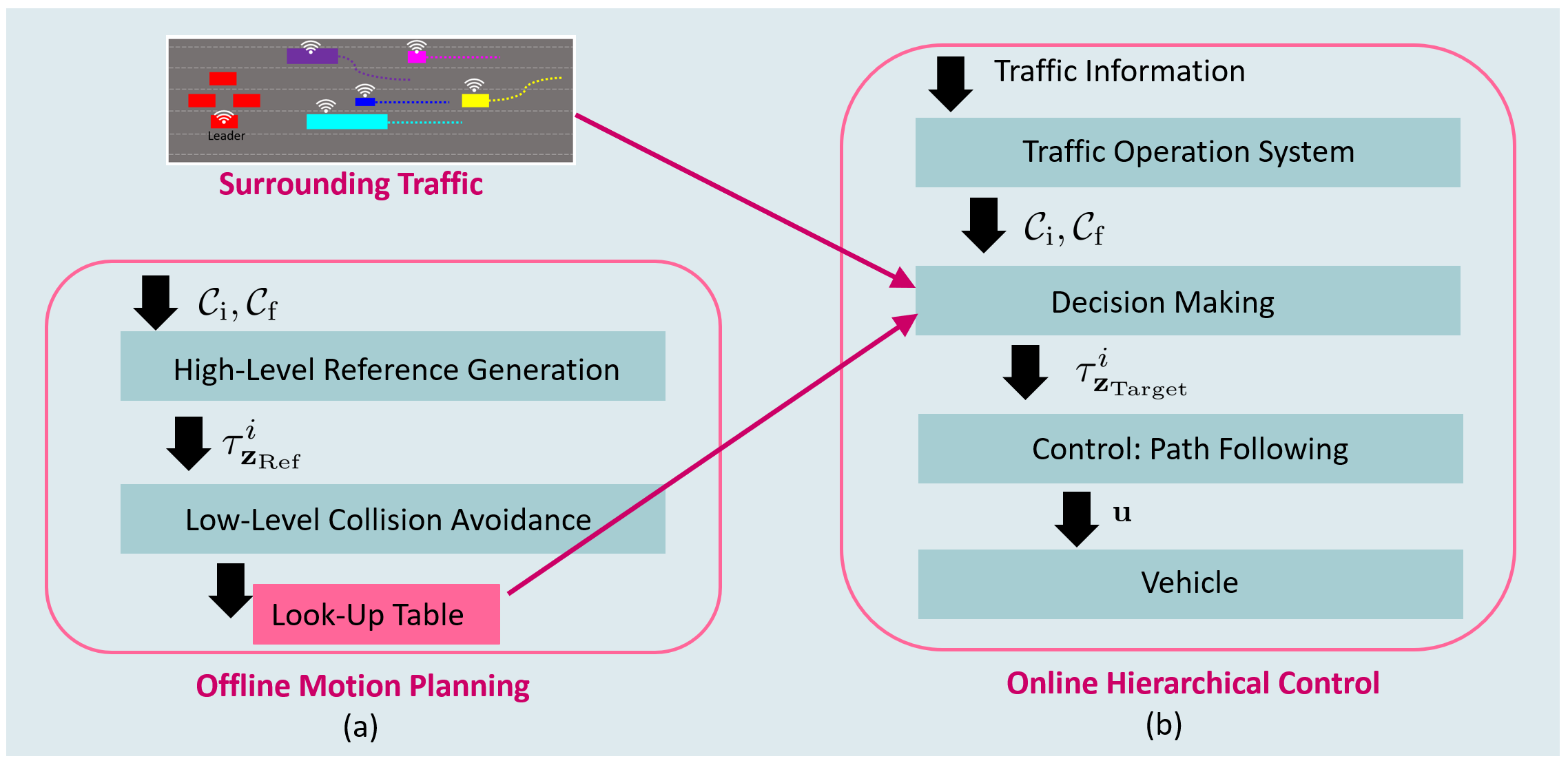}
\caption{The architecture: (a) offline motion planning system. (b) online hierarchical control system} 
\label{architecture}
\end{figure}

The proposed architecture for cooperative multi-vehicle systems consists of two main components: an offline motion planning system and an online hierarchical control system. \figref{architecture} shows the architecture. The inputs of \emph{motion planning} system are various initial $\mathcal{C}_i$ and final $\mathcal{C}_f$ configurations and the output of this system is a look-up table of precomputed safe maneuvers for transition from $\mathcal{C}_i$ to $\mathcal{C}_f$. The motion-planer uses an offline optimization-based approach for cooperative formation and reconfiguration. The \emph{online hierarchical control system} is composed of three levels: traffic operating system, decision making and path following. The \emph{traffic operating system (TOS)} monitors the traffic and determines the desired initial $\mathcal{C}_i$ and final $\mathcal{C}_f$ configurations of the platoon to improve traffic mobility and reduce road congestion. The \emph{decision-maker} receives the desired initial $\mathcal{C}_i$ and final $\mathcal{C}_f$ configurations from TOS. Also it receives future planned trajectories from the surrounding traffic $\mathcal{S}$. Based on the given desired $\mathcal{C}_i$ and $\mathcal{C}_f$ and the surrounding traffic information, the decision-maker selects a feasible transition maneuver from the look-up table to reconfigure the platoon from $\mathcal{C}_i$ to $\mathcal{C}_f$. Once the transition maneuver is selected by the decision-maker, the maneuver is executed by the \emph{path-follower} controller on each vehicle in real-time. 

The following assumptions have been made:
\begin{itemize}
\item[(A1)] The vehicles are fully autonomous and connected through vehicle-to-vehicle (V2V) and vehicle-to-cloud (V2C) communications.
\item[(A2)] All the platoon configurations $\mathcal{C}$ are selected from a pre-identified set of configurations.
\item[(A3)] The desired initial $\mathcal{C}_i$ and final $\mathcal{C}_f$ configurations are available from the topmost level of the architecture, which is the traffic operation system (TOS). The vehicles communicate with TOS via V2C communication.
\item[(A4)] Reconfiguration (transition maneuvers between initial $\mathcal{C}_i$ and final $\mathcal{C}_f$ configurations) always starts from a known (predefined) initial configuration $\mathcal{C}_i$. If the vehicles' current configuration is not identified as one of predefined configurations, the vehicles are controlled to reach the point for which $\mathcal{C}_i$ is available.  
\item[(A5)] The road is assumed to remain straight along the reconfiguration maneuver. 
\item[(A6)] Uncertainty due to communication delay or model mismatch is not considered; perfect knowledge of the states for all the vehicles is assumed.
\end{itemize}

\subsection{Motion Planning}
The motion planning is performed offline. For various identified initial $\mathcal{C}_i$ and final $\mathcal{C}_f$ platoon configurations, the transition maneuvers to reconfigure the platoon from $\mathcal{C}_i$ to $\mathcal{C}_f$ are computed by motion planner. These precomputed trajectories are stored in a look-up table to be executed by online hierarchical control system. The motion planning system has a hierarchical structure. At the high level, reference trajectories $\boldsymbol{\tau}^i_{\mathbf{z}_{\text{Ref}}}$ for each of the vehicles are generated based on initial and final configuration. These trajectories can cause collisions, which are resolved by a low level planner. At the low level, a trajectory optimization is formulated as a finite time constrained optimal control (FTCOC) problem to plan smooth, dynamically feasible and collision-free trajectories for all the vehicles in a centralized optimization problem. The motion planner incorporates the collision avoidance between the vehicles as constraints of optimization problem and obtains longitudinal $a^i$ and lateral $\delta^i$ control inputs for all the vehicles.Solving a single FTCOC optimization for the entire maneuver (until time $T$) is computationally intractable due to the large number of decision variables. Therefore, multiple FTCOC with a shorter horizon $N$ is solved, in a receding horizon fashion ($N<T$).
\subsubsection{High-Level Reference Generation}
\begin{figure}[ht]
\centering
\includegraphics[scale = 0.5]{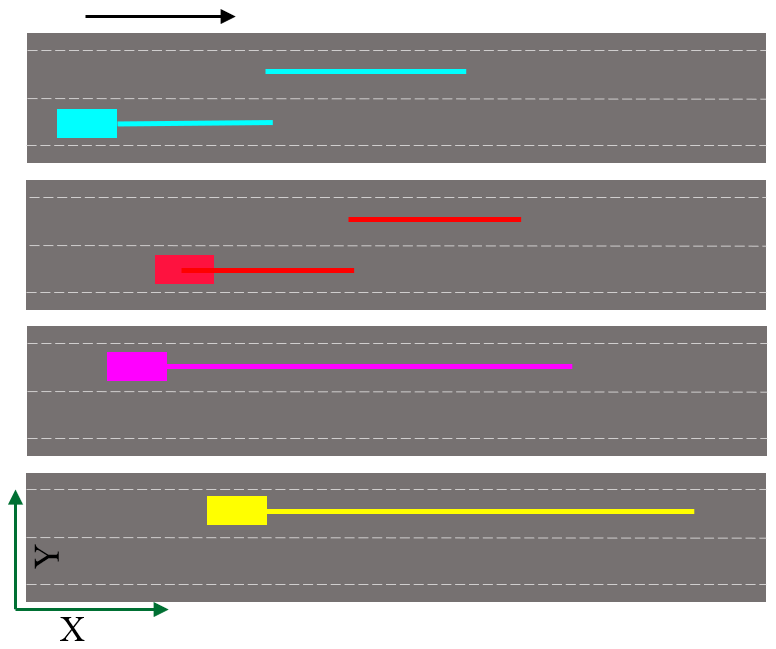}
\caption{The generated reference trajectories are shown for the example scenario of \figref{fig:reconf_example}. The parameter $\rho = 0.5$ for the blue and red vehicles.}
\label{fig:ref_generation}
\end{figure}

The reference state for $i$th vehicle is denoted as $\mathbf{z}^i_{\text{Ref}} = [x^{i}_{\text{Ref}},y^{i}_{\text{Ref}},\psi^{i}_{\text{Ref}},v^{i}_{\text{Ref}}]$. The reference state trajectory, denoted as $\boldsymbol{\tau}^i_{\mathbf{z}_{\text{Ref}}}$, is defined for the interval $[0,1,2,\hdots,T]$, from the initial time $0$ until the final maneuver time $T$ and $\boldsymbol{\tau}^i_{\mathbf{z}_{\text{Ref}}}$ = $\{\mathbf{z}^i_{\text{Ref}}(0), \mathbf{z}^i_{\text{Ref}}(1), \mathbf{z}^i_{\text{Ref}}(2),\hdots, \mathbf{z}^i_{\text{Ref}}(T)\}$.  and is computed based on initial $\mathcal{C}_{i}$ and final $\mathcal{C}_{f}$ configurations of the platoon. First, the position coordinate of all the vehicles are specified using
$g(\mathcal{C}_{i}(n_v,l,p)) = (x(0),y(0))^i \quad \forall{i} \in \mathcal{V}$, which is previously defined in \secref{secPreliminaries}. Then, the longitudinal position reference trajectory $\boldsymbol{\tau}^i_{x_{\text{Ref}}}$ = $\{x^{i}_{\text{Ref}}(0),\hdots, x^{i}_{\text{Ref}}(T)\}$ is generated using the integrator model \eqref{eq:h},
\begin{equation}\label{x_ref}
    \boldsymbol{\tau}^i_{x_{\text{Ref}}} = h(x^i(0),v_{\text{max}},T),
\end{equation}
The lateral position reference trajectory $\boldsymbol{\tau}^i_{y_{\text{Ref}}}$ is the $y$ coordinate of the road centerline for each vehicle. For the first portion of simulation $(0,\hdots,\rho T)$, $y^{i}_{\text{Ref}}$ is obtained from initial configuration $\mathcal{C}_{i}$ and the rest $((\rho T+1),\hdots,T)$ is determined by final configuration $\mathcal{C}_{f}$, $g(\mathcal{C}_{f}(n_v,l,p)) = (x^i(T),y^i(T)) \quad \forall{i} \in \mathcal{V},$
\begin{equation}\label{y_ref}
\{y^{i}_{\text{Ref}}(0),\hdots, y^{i}_{\text{Ref}}(\rho T)\}=y^i(0),
\end{equation}
\begin{equation}\label{y_ref2}
\{y^{i}_{\text{Ref}}(\rho T+1),\hdots, y^{i}_{\text{Ref}}(T)\}=y^i(T),
\end{equation}
the parameter $\rho \in (0,1)$ 
is a tuning parameter.
It is the coefficient that affects the start of the lane change. $\psi^i_{\text{Ref}}$ is zero
\begin{equation}\label{psi_ref}
   \boldsymbol{\tau}^i_{\psi_{\text{Ref}}} = \{\psi^{i}_{\text{Ref}}(0),\hdots, \psi^{i}_{\text{Ref}}(T)\} = 0,
\end{equation}
assuming the road remains straight along the maneuver and $v^{i}_{\text{Ref}}$ is set as maximum speed limit of the road or average traffic flow $v_{\text{max}}$.   
\begin{equation}\label{v_ref}
    \boldsymbol{\tau}^i_{v_{\text{Ref}}} = \{v^{i}_{\text{Ref}}(0),\hdots, v^{i}_{\text{Ref}}(T)\} = v_{\text{max}}.
\end{equation}
The reference trajectory $\boldsymbol{\tau}^i_{\mathbf{z}_{\text{Ref}}}$ for $i$th vehicle is defined using \eqref{x_ref}, \eqref{y_ref}, \eqref{psi_ref} and \eqref{v_ref}. The generated trajectory is a naive initialization that might collide with obstacles. The low-level planner ensures collision avoidance among the vehicles. \figref{fig:ref_generation} shows the generated reference trajectories for the reconfiguration scenario example \figref{fig:reconf_example}. The reference trajectories of blue and red vehicles are not straight lines, since they change their lanes. In \figref{fig:ref_generation} $\rho = 0.5$ for both blue and red vehicles.
% Note that the parameter $\rho$ determines when the lane change starts.
For example, $\rho = 0.5$ means the lane change is performed in the middle of the total duration of maneuver. For pink and yellow vehicles, $\rho$ can be any value in the interval $(0,1)$, excluding the boundaries, since pink and yellow do not change lane. 
% \begin{figure}
% \centering
% \includegraphics[scale = 0.5]{pics/reference_trajectories4.PNG}
% \caption{The reference trajectories are straight-lines that might collide with obstacles. In this example, the reference trajectory for the red car passes through the obstacle, while the others are obstacle-free.}
% \label{fig:reference_trajectories}
% \end{figure}

\subsubsection{Low-Level Collision Avoidance}
multi-vehicle motion planning problem is formulated as a centralized optimization problem that computes conflict-free trajectories for all the vehicles in the platoon simultaneously. The proposed optimization scheme uses a receding horizon fashion. At each time step it solves an optimization problem and obtains the control input based on dynamic model predictions over a time horizon and applies the first control input solution. At the next time step, the horizon is shifted forward and the procedure is repeated. The maneuvers are computed by closed-loop simulation of optimization \eqref{eq:MPC_formulation} with dynamic model \eqref{eq:kinematic_bicycle_model}. 

The objective function penalizes the deviation of each individual vehicle from the reference trajectory generated at the high level and the collision avoidance constraint is incorporated as hard constraint to guarantee safety. The optimization problem is formulated as follows
\begin{subequations}\label{eq:MPC_formulation}
\begin{align}
% \label{eq:ref_track_cost}
& \min_{\substack{\mathbf{u}^i(\cdot|t)}}
& &\nonumber{\sum_{i=1}^{N_{V}}\big(\sum_{k=t}^{t+N}||\mathbf{Q}_z(\mathbf{z}^{i}(k|t)-\mathbf{z}^{i}_\textrm{Ref}(k|t))||^2_2} \\
\label{eq:input_cost}
& & &+ \sum_{k=t}^{t+N-1}||\mathbf{Q}_u(\mathbf{u}^{i}(k|t))||^2_2 + ||\mathbf{Q}_{\Delta u}(\Delta \mathbf{u}^{i}(k|t))||^2_2\big)\\
\label{eq:dynamic_constraint}
&\textrm{subject to} & & \mathbf{z}^{i}(k+1|t) = f(\mathbf{z}^{i}(k|t),\mathbf{u}^{i}(k|t)),\\
\label{eq:initial_cond}
& & & \mathbf{z}^{i}(0|t) = \mathbf{z}^{i}(t),\\
\label{eq:states_bound}
& & & \mathbf{z}_{\min} \leq \mathbf{z}^{i}(k|t) \leq \mathbf{z}_{\max},\\
\label{eq:input_bound}
& & & \mathbf{u}_{\min} \leq \mathbf{u}^{i}(k|t) \leq \mathbf{u}_{\max},\\
\label{eq:input_change_bound}
& & & \Delta \mathbf{u}_{\min} \leq \mathbf{u}^{i}(k|t)-\mathbf{u}^{i}(k-1|t) \leq \Delta \mathbf{u}_{\max},\\
\label{collision_avoidance_constraint}
& & & \mathcal{P}(\mathbf{z}^{i}(k|t))\cap \mathcal{P}(\mathbf{z}^{j}(k|t)) = \emptyset, \quad  i \neq j\\ \nonumber
& & & \text{for all } i \in \mathcal{V},\ j \in \mathcal{N}_i,
\end{align} \end{subequations}
where $\mathbf{u}^i(\cdot|t)=\{\mathbf{u}^i(t|t),...,\mathbf{u}^i(t+N-1|t)\}$ denotes the sequence of control inputs over the planning horizon $N$ for $ith$ vehicle. The optimal solution is $U^*(t) = \{\mathbf{u}^*(t|t),...,\mathbf{u}^*(t+N-1|t)\}$, and the receding horizon control law is obtained by applying the first control input $\mathbf{u}^*(t|t).$ 

Superscript $i$ denotes the $ith$ vehicle, $N_{V}$ is the total number of vehicles in the platoon, $\mathbf{z}^{i}(k|t)$ and $\mathbf{u}^{i}(k|t)$ are the state variable and control input of $i$th vehicle at step $k$ predicted at time $t$, respectively. The above problem is a multi-objective optimization in which, the first term penalizes deviation of the states $\mathbf{z}$ from the reference state $\mathbf{z}_{\textrm{Ref}}$, the second term penalizes control input effort $\mathbf{u}$ and the third term penalizes the input rate (change of control input in two consecutive time steps) $\Delta \mathbf{u}$. The weight factors $\mathbf{Q}_z$, $\mathbf{Q}_u$ and $\mathbf{Q}_{\Delta u}$ are positive semidefinite matrices. The function $f(\cdot)$ in \eqref{eq:dynamic_constraint} represents the vehicle kinematic bicycle model (\ref{eq:kinematic_bicycle_model_discrete}), which is discretized using Euler discretization. The reference trajectory obtained from the high level planner is denoted as $\mathbf{z}^{i}_\text{Ref}$  and $\mathbf{z}_\text{min}$ and $\mathbf{z}_\text{max}$ are the state limits and $\mathbf{u}_\text{min}$ and $\mathbf{u}_\text{max}$ are the input limits. The input rate is lower bounded by $\Delta \mathbf{u}_\text{min}$ and upper bounded by $\Delta \mathbf{u}_\text{max}$. Therefore, \eqref{eq:input_change_bound} avoids heavy braking/acceleration as well as aggressive steering and enhances energy efficiency and comfort. $\mathcal{P}(\mathbf{z}^{i}(k|t))$ represents $i$th vehicle polytope as the road area occupied by the vehicle and $\mathcal{P}(\mathbf{z}^{j}(k|t))$ represents the other vehicle polytopes as moving obstacles for $i$th vehicle. The set of neighbors $\mathcal{N}_i$ is the set of all the vehicles within the platoon except $i$th vehicle and is defined as $\mathcal{N}_i = \mathcal{V} \setminus i$.
In order to guarantee collision avoidance, the vehicles are modeled as polytopic sets that not only each set has empty intersection with all the other sets, but also each set keeps a minimum distance from the other sets.  The collision avoidance between the  $i$th vehicle and all the other vehicles (neighbors) is formulated in \eqref{collision_avoidance_constraint},
where $\mathcal{P}(\mathbf{z}^{j})$ are the polytopic sets that represent all neighbor vehicles. The remainder of this section is devoted to detailed description and reformulation of the constraint \eqref{collision_avoidance_constraint}. The approach presented in \cite{03449} is used and  applied to multi-lane platoon in the next section. The underlying technical reasoning is similar and repeated here for the sake of completeness. The computed trajectories from closed simulation of optimization \eqref{eq:MPC_formulation} with dynamic model \eqref{eq:kinematic_bicycle_model} are stored in a look-up table and will be executed in real-time by a path-follower which is a feedback controller.

\subsection{Representation of the Road Area Occupied by the Vehicle}
\begin{figure}
\centering
\includegraphics[scale = 0.3]{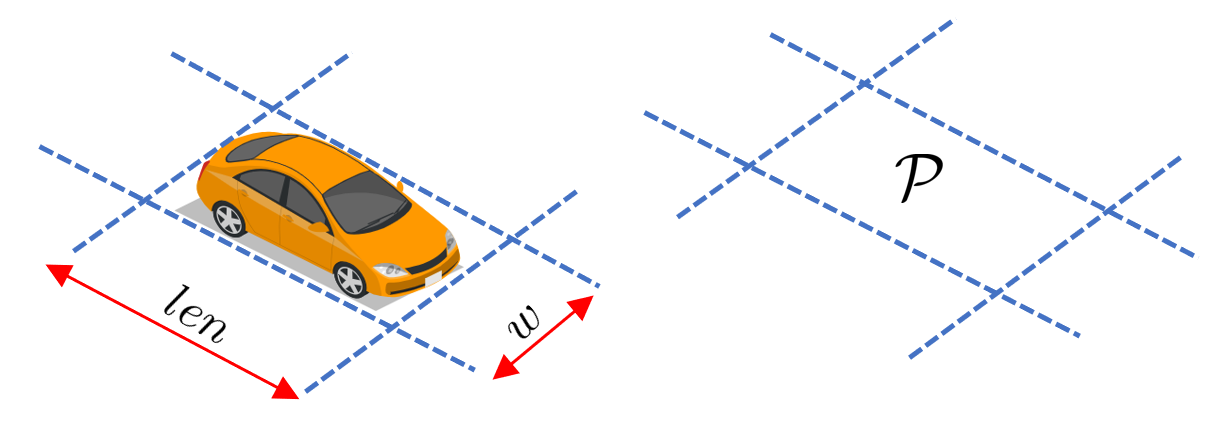}
\caption{The occupied road region is modeled as a polytopic set that undergoes affine transformations.}
\label{fig:road_region}
\end{figure}

As discussed platooning is maintaining close inter-vehicular distance within a group of vehicles. In tight platooning, both road geometry (lane width) and platoon geometry (longitudinal and lateral inter-vehicle spacing) restrict the motion of the vehicles within the platoon and results in creating a tight environment. To allow navigation at tight spaces, it is essential to model the road structure and the vehicles dimensions as exact sizes with no approximation or enlargement. The vehicle pose or the corresponding road region occupied by the vehicle is defined by a two-dimensional convex polytope $\mathcal{P}$, as seen in \figref{fig:road_region}.
The initial pose of the vehicle is represented as $\mathcal{P}_{o}$. As the vehicle travels along the road, $\mathcal{P}_{o}$ undergoes affine transformations including rotation and translation. Hence $\mathcal{P}(\mathbf{z}(k)) = \mathbf{R}(\mathbf{z}(k))\mathcal{P}_{o} + \mathbf{t}_r(\mathbf{z}(k))$, where $\mathbf{z}(k)$ represents the vehicle state at $k$th time step, $\mathcal{P}(\mathbf{z}(k))$ is the vehicle occupied region as a function of the state $\mathbf{z}(k)$, and dimensions including length $h$ and width $w$ and is defined as a set of linear inequalities. $\mathbf{R}:\mathbb{R}^{n_{z}} \rightarrow \mathbb{R}^{n\times n}$ is an orthogonal rotation matrix and $\mathbf{t}_r:\mathbb{R}^{n_{z}} \rightarrow \mathbb{R}^{n}$ is the translation vector. $n_z$ is the dimension of $\mathbf{z}$ and $n$ is two, since the transformation is occurring in two-dimensional space $\mathbb{R}^2$. The rotation matrix $\mathbf{R}(\cdot)$ is a function of the vehicle heading angle $\psi(k)$ and the translation vector $\mathbf{t}_r(\cdot)$ is a function of the longitudinal $x(k)$ and lateral $y(k)$ positions of the vehicle.
So the transformed polytope is defined as $\mathcal{P}(\mathbf{z}(k)) = \{[
p_{x},
p_{y}]^{\top}
\in \mathbb{R}^{2}
|\mathbf{A}(\mathbf{z}(k))[
p_{x},
p_{y}]^{\top}
\leq \mathbf{b}(\mathbf{z}(k))\},$ where $p_{x}$ and $p_{y}$ are the coordinates of points in two-dimensional space which are representation of the polytope. The matrix  $\mathbf{A}(\mathbf{z}(k))$ and the vector $\mathbf{b}(\mathbf{z}(k))$ are defined as
\begin{equation}\label{polytope_formula}
\begin{aligned}
&\mathbf{A}(\mathbf{z}(k)) = 
\begin{bmatrix}
\mathbf{R}(\psi(k))^{\top} \\
-\mathbf{R}(\psi(k))^{\top}
\end{bmatrix},\\
&\mathbf{b}(\mathbf{z}(k)) = [
len/2,w/2,len/2,w/2]^\top\\
&+ \mathbf{A}(\mathbf{z}(k))[
x(k),
y(k)
]^{\top},
\end{aligned}
\end{equation}
where $\mathbf{R}(\psi(k)) = 
\begin{bmatrix} 
\cos(\psi(k))& -\sin(\psi(k))\\
\sin(\psi(k)) & \cos(\psi(k)) 
\end{bmatrix}$.
The length and width of the vehicle are denoted as $len$ and $w$ , respectively, as shown in \figref{fig:road_region}. For coordination of multiple vehicles, each vehicle's occupied area is modeled as a time-varying polytope and at each time step, re-planning is performed such that no intersection occurs between the polytopic sets. 

\subsection{Collision Avoidance Reformulation}
The distance between two polytopic sets $\mathcal{P}_{1}$ and $\mathcal{P}_{2}$ is defined as \begin{equation}\label{eq:distance_polytop_primal}
\text{dist}(\mathcal{P}_{1},\mathcal{P}_{2}) =\underset{{\mathbf{x},\mathbf{y}}}{\text{min}}\{\norm{\mathbf{x}-\mathbf{y}}_{2}|\mathbf{A}_{1}x \leq \mathbf{b}_{1}, \mathbf{A}_{2} \mathbf{y} \leq b_{2}\},\\
\end{equation}
where $\mathcal{P}_{1}$ and $\mathcal{P}_{2}$ are described as $\mathbf{A}_{1}x\leq \mathbf{b}_{1}$ and $\mathbf{A}_{2}\mathbf{y}\leq \mathbf{b}_{2}$, respectively. The two sets do not intersect if $\text{dist}(\mathcal{P}_{1},\mathcal{P}_{2}) >0.$
However, for autonomous driving applications, since the vehicles must keep a minimum safe distance $d_{min}$ from each other and from the obstacles, the distance between their polytopic sets should be larger than a predefined minimum distance,  $\text{dist}(\mathcal{P}_{1},\mathcal{P}_{2}) \geq d_{min}.$ 

In the motion planning optimization problem \eqref{eq:MPC_formulation}, the collision avoidance is imposed as constraint. However, the collision avoidance formulated in \eqref{eq:distance_polytop_primal} is itself an optimization problem. Hence, an optimization problem has to be solved as the constraint of another optimization problem. To deal with this issue, as explained in \cite{03449}, the dual problem can be solved instead of the primal problem \eqref{eq:distance_polytop_primal}, based on strong duality theory. The dual problem is expressed as $\text{max}_{\boldlambda,\ \boldmu,\mathbf{s}}
\{-\mathbf{b}_{1}^{\top} \boldlambda -b_{2}^{\top} \ \boldmu:
\mathbf{A}_{1}^{\top} \boldlambda + \mathbf{s} = 0,
\mathbf{A}_{2}^{\top} \ \boldmu - \mathbf{s} = 0,
\|\mathbf{s}\| \leq 1,
\boldlambda \succeq 0, \quad \ \boldmu \succeq 0\}$, where $\boldlambda$, $\ \boldmu$ and $\mathbf{s}$ are dual variables. The optimal value of the dual problem is the distance between the two polytopes $\mathcal{P}_{1}$ and $\mathcal{P}_{2}$ and is constrained to be larger than minimum distance. Hence the constraint on dual problem optimal value is equivalent to the following feasibility problem $\{\exists \boldlambda \succeq 0, \ \boldmu \succeq 0, s:
-\mathbf{b}_{1}^{\top} \boldlambda -\mathbf{b}_{2}^{\top} \ \boldmu \geq d_\text{min},
\mathbf{A}_{1}^{\top} \boldlambda + \mathbf{s} = 0,
\mathbf{A}_{2}^{\top} \ \boldmu - \mathbf{s} = 0,
\|\mathbf{s}\| \leq 1\}.$ This reformulation can be substituted instead of collision avoidance constraint \eqref{collision_avoidance_constraint} in the motion planning optimization problem \eqref{eq:MPC_formulation}. Therefore, problem \eqref{eq:MPC_formulation} can be rewritten as 
\begin{equation}\label{eq:MPC_formulation_reformulated}
\begin{aligned}
% & \underset{{u^i(.|t)},\boldlambda_{ij}(.|t), \ \boldmu_{ij}(.|t), s_{ij}(.|t)}{\text{minimize}}
& \min_{\substack{\mathbf{u}^i(\cdot|t),\ \boldlambda_{ij}(\cdot|t),\\ \ \boldmu_{ij}(\cdot|t),\ \mathbf{s}_{ij}(\cdot|t)}}
& & \eqref{eq:input_cost} \\
&\textrm{subject to} & &
\eqref{eq:dynamic_constraint},\eqref{eq:initial_cond}, \eqref{eq:states_bound}, \eqref{eq:input_bound},\\
& & & \big(-\mathbf{b}_{i}(\mathbf{z}^{i}(k|t))^{\top} \boldlambda_{ij}(k|t) \\
& & & - \mathbf{b}_{j}(\mathbf{z}^{j}(k|t))^{\top} \ \boldmu_{ij}(k|t)\big) \geq d_\text{min},\\
& & & \mathbf{A}_{i}(\mathbf{z}^{i}(k|t))^{\top} \boldlambda_{ij}(k|t) + \mathbf{s}_{ij}(k|t) = 0,\\
& & & \mathbf{A}_{j}(\mathbf{z}^{j}(k|t))^{\top} \ \boldmu_{ij}(k|t) - \mathbf{s}_{ij}(k|t) = 0,\\
& & & \|\mathbf{s}_{ij}(k|t)\| \leq 1, -\boldlambda_{ij}(k|t) \leq 0, \\
& & & -\ \boldmu_{ij}(k|t) \leq 0, \text{for all } i \in \mathcal{V},\ j \in \mathcal{N}_i,\\
\end{aligned} \end{equation}
where $\mathbf{A}_i$ and $\mathbf{b}_i$ are functions of $\mathbf{z}^i(k|t)$ and represent the polytopic set of $i$th vehicle at step $k$ predicted at time $t$. Similarly $\mathbf{A}_j$ and $\mathbf{b}_j$ denote the polytopic set of $j$th vehicle which belongs to neighbor set $\mathcal{N}_i$. The dual variables $\boldlambda_{ij}$, $\ \boldmu_{ij}$ and $\mathbf{s}_{ij}$ are coupled through the collision avoidance constraint among vehicle $i$ and vehicle $j$. $\boldlambda_{ij}(\cdot|t)$, $\ \boldmu_{ij}(\cdot|t)$ and $\mathbf{s}_{ij}(\cdot|t)$ represent the sequence of dual variables over the optimization horizon $N$. So $\boldlambda_{ij}(\cdot|t) = \{\boldlambda_{ij}(t|t),...,\boldlambda_{ij}(t+N|t)\}$, $\ \boldmu_{ij}(\cdot|t) = \{\boldmu_{ij}(t|t),...,\ \boldmu_{ij}(t+N|t)\}$ and $\{\mathbf{s}_{ij}(\cdot|t) = \{\mathbf{s}_{ij}(t|t),...,\mathbf{s}_{ij}(t+N|t)\}$. 

One main advantage of the proposed planning method is that the required minimum distance between the vehicles $d_\text{min}$, which can be chosen as a design parameter, is always enforced during the lane change maneuvers. In theory, the trajectories can be obtained for zero $d_\text{min}$, which means the polytopic sets (cars) can move on each other boundaries. In practice, $d_\text{min}$ should be determined based on the quantification of uncertainty of physical models and stochastic measurement errors, which is one future extension of this work. 

The optimal solution of \eqref{eq:MPC_formulation_reformulated} is $U^*(t) = \{\mathbf{u}^*(t|t),...,\mathbf{u}^*(t+N-1|t)\}$, and the first control input $\mathbf{u}^*(t|t)$ is applied to the vehicle nonlinear dynamic model \eqref{eq:kinematic_bicycle_model_discrete}. Then, the initial condition is updated with the current states and the optimization \eqref{eq:MPC_formulation_reformulated} is solved again. By running forward simulations of system \eqref{eq:kinematic_bicycle_model_discrete} in closed loop with $\mathbf{u}^*(t|t)$ from the initial time $0$ to the final maneuver time $T$, one can obtain  collision-free closed-loop trajectories. Such closed-loop trajectories are represented by the state $\boldsymbol{\tau}^i_{\mathbf{z}_{\textrm{Target}}}$ trajectories. These trajectories are stored in a look-up table. The output of the motion planning system is this look-up table that captures different configurations and possible reconfigurations/transition maneuvers among them. Table \ref{tab:template} shows the structure of the look-up table. In the look-up table a set of trajectories are associated with ($\mathcal{C}_i$,$\mathcal{C}_f$) pair.

Note that for a specified pair of ($\mathcal{C}_i$,$\mathcal{C}_f$), once the high-level reference trajectory is computed, there is one optimal reconfiguration maneuver (the solution of optimization \eqref{eq:MPC_formulation_reformulated}) that transforms $\mathcal{C}_i$ to $\mathcal{C}_f$. However, the high-level reference trajectory computed by \eqref{y_ref} and \eqref{y_ref2} is parameterized by $\rho \in (0,1)$ and different choices of $\rho$ result in different reference trajectories and consequently various reconfiguration maneuvers. In practice, several different values of $\rho$ can be chosen, for example $\rho = \{0.1,0.2,0.3$,$0.4,0.5,0.6$,$0.7,0.8,0.9\}$ and the look-up table can be computed with these $\rho$ values. So the family of reconfiguration maneuvers from $\mathcal{C}_i$ to $\mathcal{C}_f$ is computed for these specified $\rho$ values. The number of reconfiguration trajectories for a specific ($\mathcal{C}_i$,$\mathcal{C}_f$) pair is restricted to $M$ numbers due to limited memory storage. The look-up table for specified $\rho$ values is shown in Table \ref{tab:look-up_tab_with_rho}. The nonlinear optimization \eqref{eq:MPC_formulation_reformulated} is not persistently feasible. The infeasible solutions are discarded and not included in the look-up table. Note that the motion-planner avoids collisions among the vehicles within the platoon $(i \in \mathcal{V})$. However, collision avoidance with surrounding traffic (vehicles outside the platoon), should be considered by decision-maker, as explained in \secref{decision-maker}. 

\begin{table}[]
\caption{Look-up table structure (Parameterized by $\rho$).}\label{tab:template}
\centering
\begin{tabular}{|c|l|}
\hline
\textbf{Reconfiguration} & \textbf{Trajectories $\forall i \in$ $\mathcal{V}$} \\
\hline
\multirow{2}{*}{$(\mathcal{C}_i, \mathcal{C}_f)$} &  $\boldsymbol{\tau}^i_{z_{\textrm{Target}}}(\rho^i)$\\
& $\rho^i \in (0,1)$\\
\hline
\end{tabular}
\end{table}

\begin{table}[]
\caption{Look-up table structure (for specified $\rho$ values).}\label{tab:look-up_tab_with_rho}
\centering
\begin{tabular}{|c|c|c|}
\hline
\textbf{Reconfiguration} & \textbf{Index} &\textbf{Trajectories} $\forall i \in$ $\mathcal{V}$ \\
\hline
\multirow{6}{*}{$(\mathcal{C}_i, \mathcal{C}_f)$} &  1& $\boldsymbol{\tau}^i_{z_{\textrm{Target}}}(\rho^i = 0.1)$\\
& 2 & $\boldsymbol{\tau}^i_{z_{\textrm{Target}}}(\rho^i = 0.2)$\\
& \vdots & \vdots \\
& $M$ & $\boldsymbol{\tau}^i_{z_{\textrm{Target}}}(\rho^i = 0.9)$\\
\hline
\end{tabular}
\end{table}

\subsection{Traffic Operation System}
The traffic operation system (TOS) is the topmost level of online hierarchical control system and operates in the cloud. TOS determines the desired initial $\mathcal{C}_i$ and final $\mathcal{C}_f$ configurations based on the road traffic information. This level selects the desired platoon configurations among all the pre-identified configurations within the set $\mathbb{C}$, in such a way to improve traffic mobility and to reduce traffic congestion. The vehicles communicate with this level via V2C communication and receive $\mathcal{C}_i$ and $\mathcal{C}_f$. Determining the optimal desired configuration can be done with rule-based method and its discussion is out of the scope of this paper. Therefore, in this paper, based on assumption (A3), it is assumed that the initial $\mathcal{C}_i$ and final $\mathcal{C}_f$ configurations are already determined by TOS and are given to the vehicles.         
% A general finite state machine is introduced and illustrated in \figref{fig:state_machine}(a) and characterized by $F(\bar{\mathcal{C}_{i}},\bar{\mathcal{C}_{f}}, e_t, T_r, e_f)$, where

%     \item  $\bar{\mathcal{C}_{i}}, \bar{\mathcal{C}_{f}}$ are discrete states that are instances of platoon configurations. $\bar{\mathcal{C}_{i}}$ is the initial configuration and $\bar{\mathcal{C}_{f}}$ is the final configuration.
%     \item $e_t$ is the event-triggered switch which is defined as a binary function$f_{b}:(u,G,\text{exo.vars.})\rightarrow(\{0,1\}),$
%     \begin{equation}\label{binary_function}
%         e_t = f_{b}(u,G,\text{exo.vars.}), 
%     \end{equation}
%     where $u$ is the input, $G$ is the state guard and $\text{exo.vars}$ are the exogenous variables.
%     \item $e_f$ presents the final event in which the platoon steady state is achieved and  configuration $\bar{\mathcal{C}_{f}}$ is realized. 
% To be more specific, examples of the above quantities are provided. Input $u$ in \eqref{binary_function} is the signal contains the traffic information which can be obtained by the following examples

\subsection{Decision-Making}\label{decision-maker}

The decision-making system runs on an individual vehicle in the platoon. The front-most vehicle in the reference lane, is chosen as platoon leader on which the decision-maker operates. The decision-maker receives three types of information: 1) the desired initial $\mathcal{C}_i$ and final $\mathcal{C}_f$ configurations from the TOS obtained via cloud; 2) look-up table computed by motion-planner pre-stored on the vehicle; 3) the surrounding (outside of platoon) vehicles real-time information obtained via V2V communication. This information includes the vehicles current states and future plans over the horizon that is equal or longer than the reconfiguration maneuver duration $T$. The decision-maker makes use of this information to check whether the desired reconfiguration maneuver planned by TOS is feasible or not.
\begin{figure}
\centering
\includegraphics[width=\linewidth]{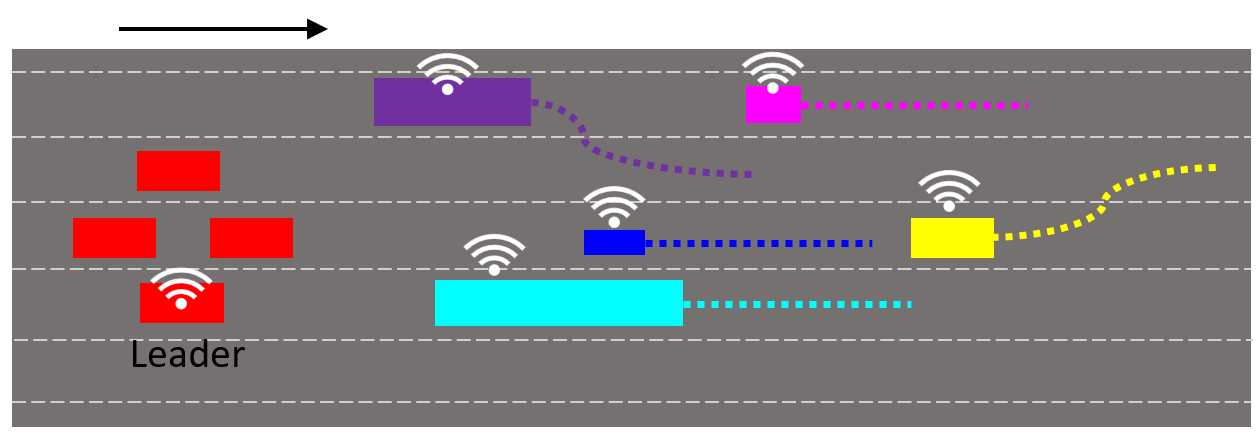}
\caption{The red vehicles are within the platoon $i \in \mathcal{V}$ and the rest of vehicles with different colors are the surrounding traffic (outside platoon) vehicles $q \in \mathcal{S}$. The outside platoon vehicles share their future planned trajectories $\boldsymbol{\tau}^q_\mathbf{z}$, with the platoon leader (decision-maker) via V2V communication. The dotted lines illustrate the planned trajectories of surrounding traffic vehicles.}
\label{fig:CAVs_shared_plans}
\end{figure}

Based on assumption (A1),
% all the vehicles are assumed to be autonomous and equipped with V2V communication, so
the surrounding vehicles (outside of platoon) can communicate with decision-maker and share their future planned trajectories with it, as shown in \figref{fig:CAVs_shared_plans}. The surrounding traffic vehicles are defined as the set $\mathcal{S}$ and their future planned trajectories are denoted as $\boldsymbol{\tau}^q_\mathbf{z}$. Superscript $q$ is index of the outside platoon vehicle. Decision-maker is responsible to ensure the planned desired reconfiguration from $\mathcal{C}_{i}$ to $\mathcal{C}_{f}$ is collision-free with respect to outside platoon traffic. To do a collision check between the platoon vehicles $i \in \mathcal{V}$ and surrounding traffic (outside platoon vehicles) $q \in \mathcal{S}$, their trajectories must be compared. Therefore, platoon reconfiguration trajectories $\boldsymbol{\tau}^i_{z{\textrm{Target}}}$ and the surrounding vehicles (not in platoon) future planned trajectories $\boldsymbol{\tau}^q_{\mathbf{z}}$ should be checked for collision at each time instant $t$. The positions $x(t)$, $y(t)$ and heading $\psi(t)$ are included in these trajectories. In addition the shared information of the surrounding vehicles include the vehicles dimensions including length $len$ and width $w$.

Given two trajectories $\boldsymbol{\tau}^1_{\mathbf{z}} = \{\mathbf{z}^1(0), \hdots, \mathbf{z}^1(T)\}$ and $\boldsymbol{\tau}^2_\mathbf{z} = \{\mathbf{z}^2(0), \hdots, \mathbf{z}^2(T)\}$ associated with vehicles $1$ and $2$, respectively, the vehicles polytopic representations $\eqref{polytope_formula}$ at each time instant can be used to check whether the vehicles collide or not. The Algorithm \ref{alg:collision-check} explains the collision-check procedure. Step \textcircled{\footnotesize{4}} computes the polytopic representation of the vehicle $1$, $\mathbf{A}_1(t)$, $\mathbf{b}_1(t)$ and the polytopic representation of the vehicle $2$, $\mathbf{A}_2(t)$, $\mathbf{b}_2(t)$, for all the time steps $t \in {0,1, \hdots, T}$ in parallel. To do so, the algorithm uses \eqref{polytope_formula}. Since $\boldsymbol{\tau}^1_z$ includes the information $\mathbf{z}^1(t) = [x^1(t),y^1(t),\psi^1(t),v^1(t)]$, by substituting $x^1(t),y^1(t),\psi^1(t)$ and by including the vehicle dimensions, length $len^1$ and width $w^1$ in \eqref{polytope_formula}, the polytopic representation $\mathbf{A}_1(t)$, $\mathbf{b}_1(t)$ can be computed. The same procedure is repeated to find $\mathbf{A}_2(t)$, $\mathbf{b}_2(t)$. Step \textcircled{\footnotesize{5}} uses \eqref{eq:distance_polytop_primal} to compute the distance, $dist(t)$ between the two polytopic representations (vehicles). When the distance is less than an acceptable safe minimum distance $d_{\text{min}}$, the collision has occurred and the algorithm outputs true collision flag, otherwise it outputs false collision flag. Note that the problem \eqref{eq:distance_polytop_primal} is a simple convex problem which is computationally cheap and suitable for real-time implementation.  

\begin{algorithm}[]
\caption{\small Collision Check Algorithm}
    \begin{algorithmic}[1]
        \State {\textbf{Inputs:}  $\boldsymbol{\tau}^1_z, \boldsymbol{\tau}^2_z$, and vehicles dimensions: $len^1,w^1$ and $len^2,w^2$}
        \State \textbf{Output:} collision flag { \Comment{($\boldsymbol{\tau}^1_z$ and $\boldsymbol{\tau}^2_z$ collide if flag=True, and $\boldsymbol{\tau}^1_z$ and $\boldsymbol{\tau}^2_z$ do not collide if flag=False.)}}
        \ForAll{ $t \in \{0,1, \hdots, T\}$} { in parallel}
        \State {compute $\mathbf{A}_1(t)$, $\mathbf{b}_1(t)$ and $\mathbf{A}_2(t)$, $\mathbf{b}_2(t)$, the polytopic representation of vehicle $1$ and vehicle $2$, by using equation \eqref{polytope_formula}.}
        \State{compute the distance $dist(t)$ between the two vehicles (polytopes), by solving problem \eqref{eq:distance_polytop_primal}.}
        \If {$dist(t)\geq d_{\text{min}}$} 
        \State{collision flag = False}
        \Else  
        \State {collision flag = True, and go to step 12}
        \EndIf
        \EndFor\\
        \Return{collision flag}
    \end{algorithmic}
    \label{alg:collision-check}
\end{algorithm}

The Algorithm \ref{alg:decision_making} explains the decision making process. At step \textcircled{\footnotesize{3}}, the decision-maker queries the pre-stored look-up table to get the family of trajectories associated with the pair $(\mathcal{C}_{i}, \mathcal{C}_{f})$. Step \textcircled{\footnotesize{5}} uses Algorithm \ref{alg:collision-check} to check the collision between the selected platoon reconfiguration maneuver $\boldsymbol{\tau}_{\mathbf{z}_{\textrm{Target}}}$ and the future planned trajectories $\boldsymbol{\tau}^q_{\mathbf{z}}$ shared by surrounding (not in platoon) vehicles. The search over the family of trajectories in the look-up table is continued until finding a feasible platoon reconfiguration maneuver which has no conflict with the surrounding (not in platoon) vehicles. If the search finishes and no conflict-free reconfiguration maneuver is found, the decision-maker informs TOS that the current reconfiguration plan is infeasible. 
So, the plan is canceled and the vehicles will move forward with the current configuration. The decision-maker waits for the TOS to plan a new reconfiguration for the future time.
If a feasible maneuver is found, it will be broadcasted through V2V network to all the vehicles and each vehicle executes its own trajectory in real-time via low-level path-following controller.

\begin{algorithm}[] 
\caption{\small Decision-Making Algorithm}
    \begin{algorithmic}[1]
        \State {\textbf{Inputs:}  $\mathcal{C}_i, \mathcal{C}_f$, look-up table, $\boldsymbol{\tau}^q_{\mathbf{z}} \quad \forall{q \in \mathcal{S}}$.}
        \State {\textbf{Output:} $\boldsymbol{\tau}_{\mathbf{z}_{\textrm{Target}}}(j)$ or infeasible flag.}
        \State {query the look-up table to find the family of trajectories $\boldsymbol{\tau}_{\mathbf{z}_\textrm{Target}}$ associated with the pair $(\mathcal{C}_i, \mathcal{C}_f)$.}
        \For {j = 1 to M}
        \State{check the collision between $\boldsymbol{\tau}_{{\mathbf{z}}_\textrm{Target}}(j)$ and $\boldsymbol{\tau}^q_{\mathbf{z}}$ by running the Algorithm \ref{alg:collision-check}.}
        \If {collision flag = False}
        \State{\Return{$\boldsymbol{\tau}_{\mathbf{z}_\textrm{Target}}(j)$}}
        \EndIf
        \EndFor\\
        \Return {infeasible flag}
    \end{algorithmic}
    \label{alg:decision_making}
\end{algorithm}

\subsection{Path-Following}
The path-following controller on each vehicle $i$ executes $ith$ vehicle corresponding maneuver and operates in real time. The desired maneuver (communicated by decision-maker) is represented by the state trajectory $\boldsymbol{\tau}_{\mathbf{z}_{\text{Target}}}$ = $\{\mathbf{z}_\textrm{Target}(0), \hdots $, $\mathbf{z}_\textrm{Target}(T)\}$. The path-follower is designed using model predictive control (MPC) as follows
\begin{subequations}\label{eq:path_follower}
\begin{align}
& \min_{\substack{\mathbf{u}(\cdot|t)}}
& &\nonumber{\bigg(\sum_{k=t}^{t+N}||\mathbf{Q}_z^{pf}(\mathbf{z}(k|t)-\mathbf{z}_\textrm{Target}(k|t))||^2_2} \\
& & &+ \nonumber{\sum_{k=t}^{t+N-1}\big(||\mathbf{Q}_{u2}^{pf}(\mathbf{u}(k|t))||^2_2}\\
& & &+ \nonumber{||\mathbf{Q}_{\Delta u}^{pf}(\Delta \mathbf{u}(k|t))||^2_2\big)\bigg)}\\
\label{dynamic_path_following}
&\textrm{subject to} & & \mathbf{z}(k+1|t) = f(\mathbf{z}(k|t),\mathbf{u}(k|t)),\\
& & & \mathbf{z}(0|t) = \mathbf{z}(t),\\
& & & \mathbf{z}_{\min} \leq \mathbf{z}(k|t) \leq \mathbf{z}_{\max},\\
& & & \mathbf{u}_{\min} \leq \mathbf{u}(k|t) \leq \mathbf{u}_{\max},\\
\label{input_rate_path_following}
& & & \Delta \mathbf{u}_{\min} \leq \mathbf{u}(k|t)-\mathbf{u}(k-1|t) \leq \Delta \mathbf{u}_{\max},
\end{align} \end{subequations}
where the notations are similar to the notations in problem \eqref{eq:MPC_formulation}. The superscript $i$ is removed, because each car independently runs the path-following controller. The first term of the objective penalizes the state deviation from the target state trajectory $\boldsymbol{\tau}_{\mathbf{z}_\textrm{Target}}$, the second and third terms penalize control input effort and input rate, respectively. The weight factors, $\mathbf{Q}_z^{pf}$, $\mathbf{Q}_{u1}^{pf}$, $\mathbf{Q}_{u2}^{pf}$ and $\mathbf{Q}_{\Delta u}^{pf}$ are super-scripted by $pf$ to be distinguished from the weight factors in problem \eqref{eq:MPC_formulation}. These weight factors should be tuned to achieve high tracking performance. The constraints \eqref{dynamic_path_following}-\eqref{input_rate_path_following} are the same as the constraints \eqref{eq:dynamic_constraint}-\eqref{eq:input_change_bound} in problem \eqref{eq:MPC_formulation}.

% In $\mathcal{C}_{i}$ and $\mathcal{C}_{f}$ modes the platoon is at steady state, so for these modes the controllers are lane keeping and Adaptive Cruise Control (ACC). In $T_r$ mode the controller is a path-follower.

\subsection{Configuration Design Heuristics}
\begin{figure}
\centering
\includegraphics[scale = 0.35]{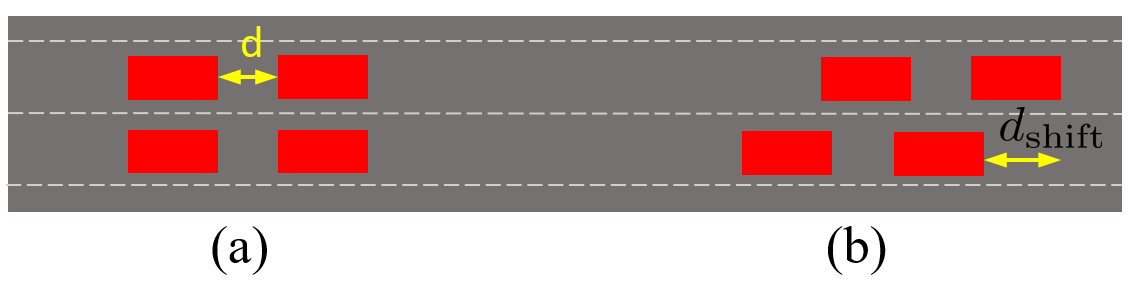}
\caption{(a) The configuration without shifting distance in adjacent lanes. (b) The configuration with shifting distance.}
\label{fig:shifted_distance}
\end{figure}

The two main factors that should be considered in configuration design are 1) inter-vehicle longitudinal spacing in one lane $d$ and 2) the shifting distance in two adjacent lanes $d_{\text{shift}}$. The small inter-vehicle longitudinal gap reduces air drag, contributes to energy saving and improves traffic throughput, as discussed earlier. In addition, the small gap prevents the surrounding traffic (outside platoon vehicles) to cut-in between the platoon vehicles. On the other hand the longitudinal spacing should be large enough to ensure safety and robustness to uncertainties. 
Furthermore, a shifting distance between two adjacent lane facilitates lane-change maneuvers. \figref{fig:shifted_distance}(a) shows a platoon configuration with no shifting distance in adjacent lanes. \figref{fig:shifted_distance}(b) shows a configuration with shifting distance in adjacent lanes. The platoon configuration (b) is more flexible for reconfiguration compared to the platoon configuration (a). Choosing the optimal values of $d$ and $d_{\text{shift}}$ should be done using experimental data and is another extension of this work.

\section{Formation as Sequence of Motion Primitives}
\label{secSimpleExample}

An alternative approach for the proposed optimization-based motion planning is behavior-based planning. In this section, a behavior-based planning using sequence of motion primitives \cite{736776} is reviewed. This behavior-based approach is used to benchmark the proposed optimization-based planning against. As described in \secref{secLiteratureReview}, among all the existing methods, the behavior-based approach which uses a sequence of motion primitives is more suitable for formation of multi-lane platoons. In robotics applications, a complex dynamical task is achieved by synthesizing a sequence of motion primitives. In a similar way, achieving the desired platoon formation requires that a sequence of motion primitives to be performed by each single vehicle in the platoon. This method is considered as a baseline and the proposed optimization-based motion planning approach is compared with this behavior-based method using a simple example scenario in \secref{secResults} and the advantages of the proposed approach are discussed. 

For each motion primitive a number of parameters have to chosen. The examples of parameterized motion primitives for a single car in multi-lane formation are  
\begin{itemize}
    \item slow down: parameterized by desired speed and desired deceleration),
    \item cruise control (CC): parameterized by desired speed
    \item lane change: parameterized by lane index, desired acceleration or deceleration),
    \item adaptive cruise control (ACC): parameterized by the front's car velocity and the desired inter-vehicle distance.
\end{itemize}

Planning sequence of motion primitives for each vehicle in the platoon to achieve a certain formation is hard to formulate and analyze mathematically. In this method, the system of vehicles is modeled as a hybrid system with various motion primitives as discrete modes and the transition maneuvers between them as continuous dynamics. To plan a sequence of motion primitives a mixed-integer program (MIP) has to be solved, where different types of motion primitives are integer decision variables and the vehicles' states are the continuous decision variables. However, MIPs are in general difficult to solve. An alternative common approach is to obtain the sequence of motion primitives according to a rule-based approach and then execute each motion primitives using the individual controllers for each primitives. Since the study of behavior-based approach is not the focus of this paper, the problem is simplified and the sequence of motion primitives for each vehicle are assumed to be already determined based on some rules. Given the sequence of primitives, the controllers are designed to execute them. All the controllers are designed, using MPC scheme such that the reference tracking cost is minimized while respecting vehicle dynamics and input and state limits. To keep the brevity of the paper, the controllers' mathematical formulations are not discussed here, but detailed description can be found in the authors' previous works. For example, an MPC cruise controller (CC) discussed in \cite{8619538} is designed to execute following a desired velocity. Also, an adaptive cruise control (ACC) is designed to maintain a proper distance from the front car and follow the front car's velocity, using the MPC formulation described in \cite{8814928}. "Lane change" is achieved by changing the center of lanes as reference. In \secref{secResults}, these controllers are used to execute the given motion primitives for a simple example scenario for multi-vehicle formation.
%\begin{remark}
%The goal is not to show "superiority" of one method versus the %other, but rather to highlight difference. 
%\end{remark}

\section{Numerical Results}
\label{secResults}
Three simulation scenarios are conducted to verify the effectiveness of the proposed motion planning algorithm. The simulations are conducted in MATLAB, the optimization problem is modeled using YALMIP and the nonlinear optimization is solved using IPOPT. The results are reported for three cases: a) platoon formation and re-configuration, b) obstacle avoidance, and c) comparison with behavior-based approach. The vehicle dimensions are chosen as $4.5$m length and $1.8$m width. The road width is chosen as $3.7$m, which is the highway lane width standard at the United States.  The control input limits are chosen as realistic physical limits of actual passenger vehicle. The acceleration input lower and upper bounds are chosen as $-4$m/s${}^2$ and $4$m/s${}^2$, respectively and its change is limited to $-1$m/s${}^2$ and $1$m/s${}^2$. The steering input lower and upper bounds are chosen as $-0.3$rad and $0.3$rad and its change is limited to $0.2$rad/s. At each iteration the optimization problem \eqref{eq:MPC_formulation_reformulated} is solved and the first control input is applied to the vehicle kinematic model \eqref{eq:kinematic_bicycle_model} for all the vehicles. Then the horizon is shifted and same procedure is repeated for the next step. For all the three scenarios the simulation results are presented as top view snapshots, as well as a series of state and action plots. The vehicles colors of the snapshots and plots are matched. The video for formation reconfiguration and obstacle avoidance scenarios is available online at this link \url{https://github.com/RoyaFiroozi/Centralized-Planning}.    

\subsection{Platoon Re-Configuration}
\begin{figure*} 
  \centering
  \includegraphics[width = 0.7\linewidth]{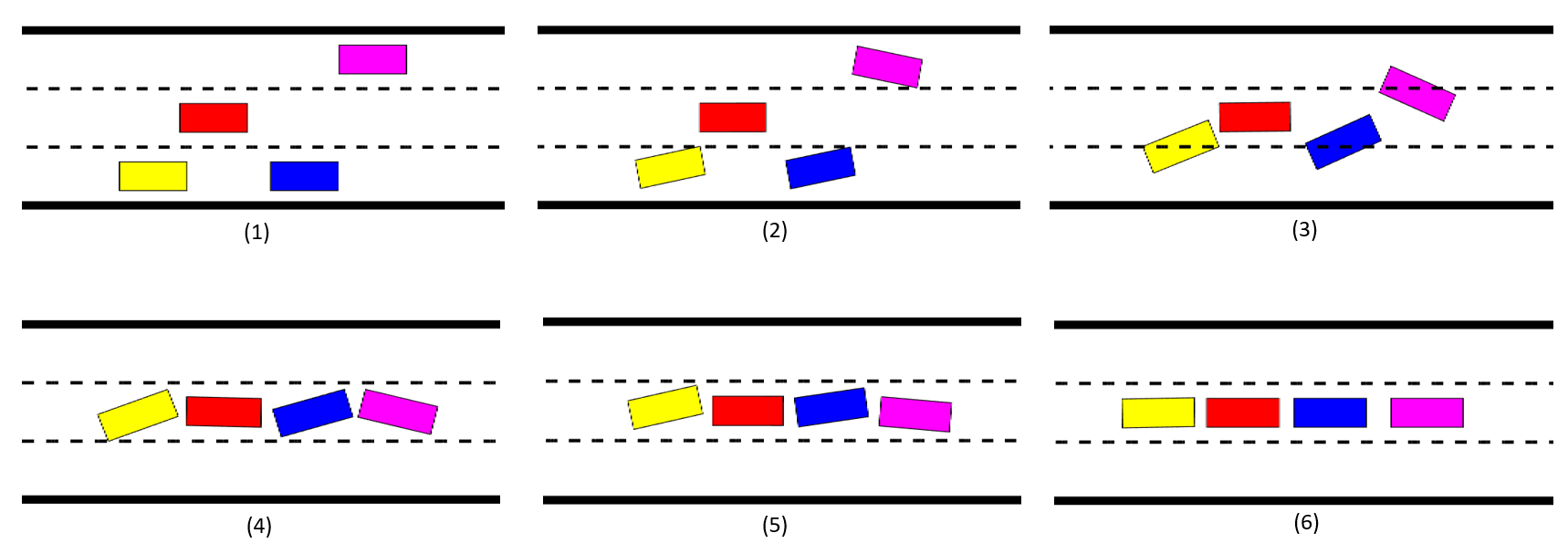}
  \centering
  \includegraphics[width = \linewidth]{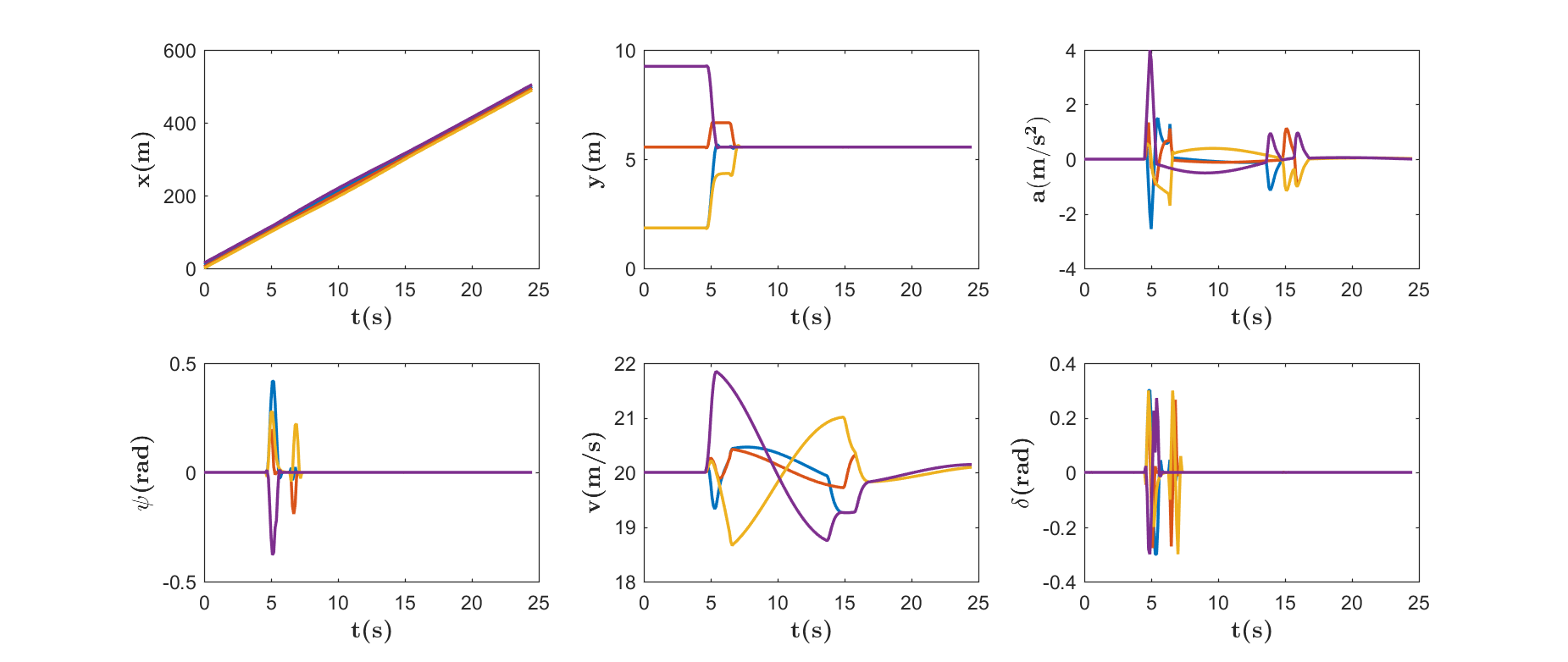}
  \caption{\textbf{Top:} Platoon reshapes from multi-lane configuration into single-lane configuration. Four vehicles moving in three different lanes merge in one lane. Step (1) shows four vehicles moving to the right in three different lanes at steady state. Steps (2) to (5) show the merging maneuver and finally step (6) demonstrates single-lane platoon configuration as another steady state of the platoon. \textbf{Bottom:} The vehicles' states and actions in a merging maneuver are presented. The colors of all the plots are matched with the color of the vehicles in top view snapshots.}
  \label{fig:snapshot_4vehivle}
\end{figure*}

In this scenario the platoon formation is alternating between two different configurations, as seen in \figref{fig:snapshot_4vehivle}. The platoon of four vehicles is moving in a two-dimensional configuration. The vehicles are moving in three different lanes and the platoon reshape into one-dimensional configuration and all the vehicles merge into one lane. The initial configuration is $\mathcal{C}_i(2,[1,1,1],p_i)$, with $p_i=[0, 5.5;6, 0;-4.5, 0]$ (matrix rows are separated by semicolons). The final configuration is $\mathcal{C}_f(4,[0,1,0]),p_f$, with $p_f=[0,0,0,0;0,0.3,0.3,0.3;0,0,0,0]$. The initial longitudinal coordinates for all the four vehicles are $[x^1(0),x^2(0),x^3(0),x^4(0)]=[10.5,4.5,0.5,15]$ and the initial lateral coordinates are $[y^1(0),y^2(0),y^3(0),y^4(0)]=[1.85,5.55,1.85,9.25]$. $d_\text{min}$ is chosen as $0.3$m, the horizon $N$ is 5, sampling time $\Delta t$ is $0.2$s, simulation time $T$ is $120$, $\rho$ is $0.25$ and $v_{\text{max}}$ is $20$m/s. \figref{fig:snapshot_4vehivle} represents the vehicles' states and actions. The plots show the transient behavior between the two modes or configurations. The longitudinal and lateral coordinates $x$ and $y$, as well as heading angle $\psi$ and velocity $v$ for all the vehicles are shown in different colors which are matched with the colors in \figref{fig:snapshot_4vehivle}. The control actions $a$ and $\delta$ are also illustrated for all the vehicles. As seen the platoon reaches its steady state at final configuration after about $25$ seconds.   

% \begin{figure*}
% \makebox[\linewidth][c]{20
% \begin{subfigure}[a]{1\paperwidth}
%     \centering
%   \includegraphics[width = 0.6\paperwidth]{pics/1D_platoon_formation2.PNG}
%   \caption{}
%   \label{fig:snapshot_4vehivle}
%   \end{subfigure}
% }
% \makebox[\linewidth][c]{20
%     \begin{subfigure}[b]{1\paperwidth}
%     \centering
%     \includegraphics[width=0.8\paperwidth]{pics/plot_results_formation_v20.png}
%   \caption{}
%   \label{fig:results_platoon_reconfiguration}
%   \end{subfigure}
%   }
%   \caption{(a) Platoon reshapes from 2D into 1D configuration. Four vehicles moving in three different lanes merge in one lane. Step (1) shows four vehicles moving to the right in three different lanes. This 2D configuration represents a steady state of the platoon. Steps (2) to (5) show the merging maneuver and finally step (6) demonstrates 1D platoon configuration as another steady state of the platoon. (b) The vehicles' states and actions in a merging maneuver are presented. The plots demonstrate the transient between two configuration (steady state). The colors of all the plots are matched with the color of the vehicles in top view snapshots.}
% \end{figure*}

\subsection{Obstacle Avoidance}
\begin{figure}
\makebox[\columnwidth][c]{
\begin{subfigure}[]{1\columnwidth}
    \centering
    \includegraphics[width = \columnwidth]{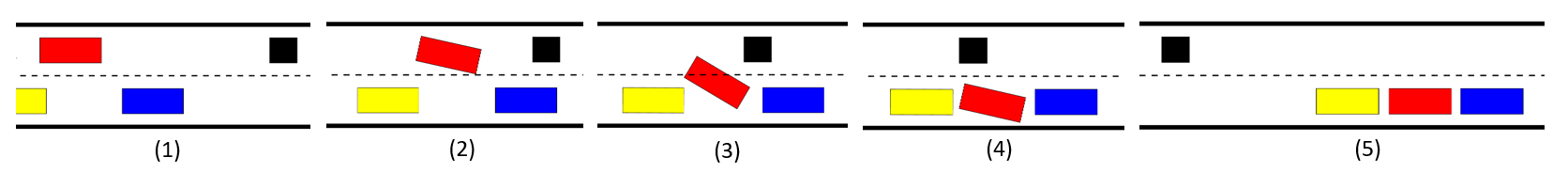}
    \caption{}
    \label{fig:obstacle_avoidance_snapshot}
\end{subfigure}
}
\makebox[\columnwidth][c]{
\begin{subfigure}[]{1.2\columnwidth}
    \centering
    \includegraphics[width =  1\columnwidth]{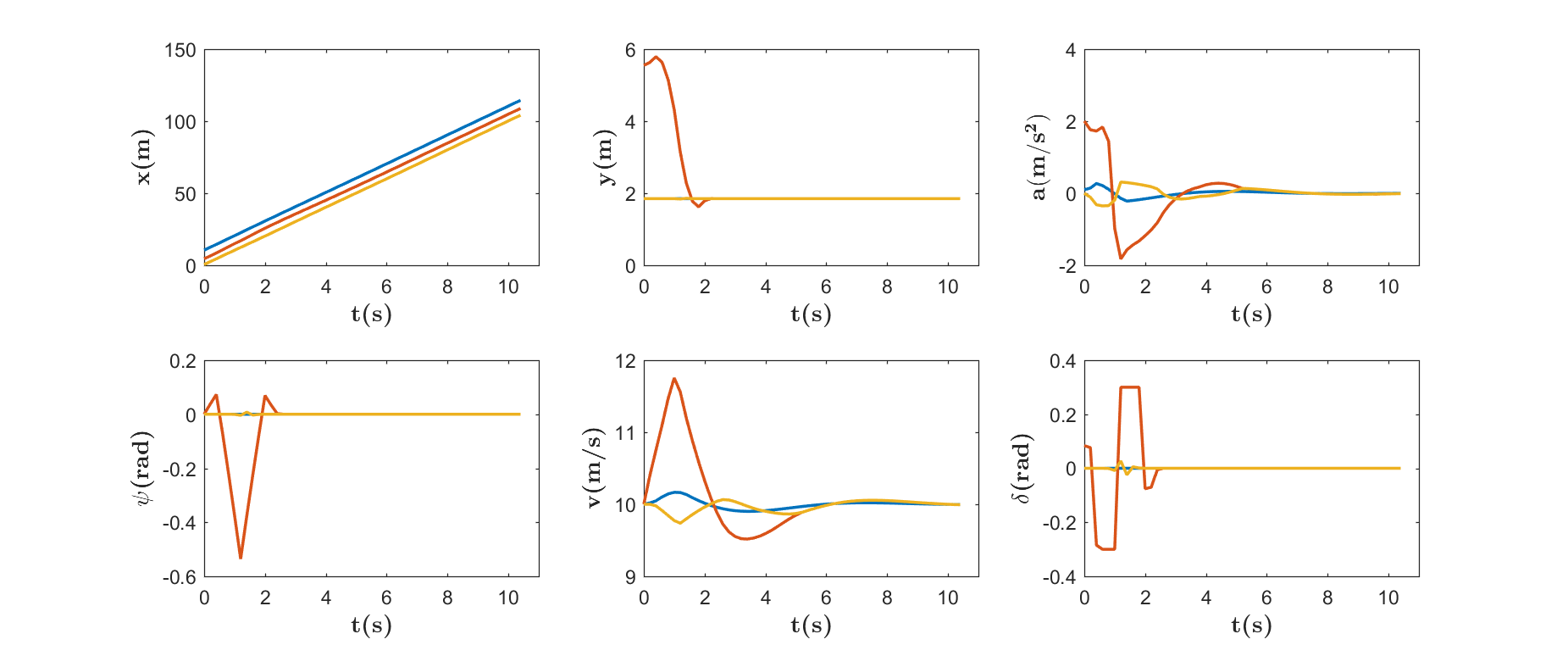}
    \caption{}
    \label{fig:results_collision_avoidance}
\end{subfigure}
}
\caption{(a) A static obstacle (black object) is detected in the red vehicle lane, the yellow and blue vehicles make gap for the red to merge into their lane. Vehicles are travelling in 2D configuration are flexible and are able to reshape in case of presence of obstacle in one lane. (b) The plots correspond to the snapshots and represents the vehicles' states and actions during the simulation. As seen, the steady state is achieved and 1D configuration is formed.}
\end{figure}

In obstacle avoidance scenario multiple vehicles are traveling together in a multi-lane platoon formation and once an obstacle is detected in the left Lane, the TOS selects reconfiguration to a single-lane configuration in the right lane. 
The vehicles in the other lane make enough gap to facilitate safe and smooth lane changing and merging for the vehicles in the lane with obstacle. \figref{fig:obstacle_avoidance_snapshot} shows the top view snapshots for obstacle avoidance simulation. The red vehicle has to change lane because a static obstacle (black object) has been detected on its lane. The yellow and blue vehicles make gap for the red vehicle to merge into their lane. The obstacle is modeled as a polytopic set and the obstacle avoidance constraints are introduced. The initial longitudinal coordinates for all the three vehicles are $[x^1(0),x^2(0),x^3(0)]=[10.5,4.5,0.5]$ and the initial lateral coordinates are $[y^1(0),y^2(0),y^3(0)]=[1.85,5.55,1.85]$. $d_\text{min}$ is chosen as $0.2$m, the horizon $N$ is 8, sampling time $\Delta t$ is $0.1$s, simulation time $T$ is $100$, $\rho$ is $0.25$ and $v_{\text{max}}$ is $10$m/s. The vehicles' states and actions are shown for in \figref{fig:results_collision_avoidance}. As seen, the steady state is achieved and 1D platoon is formed after about $10$ seconds.   

% \begin{figure*}
% \makebox[\linewidth][c]{20
% \begin{subfigure}[a]{1\paperwidth}
%     \centering
%     \includegraphics[width = 0.85\paperwidth]{pics/obstacle_avoidance_snapshot.PNG}
%     \caption{}
%     \label{fig:obstacle_avoidance_snapshot}
% \end{subfigure}
% }
% \makebox[\linewidth][c]{20
% \begin{subfigure}[b]{1\paperwidth}
%     \centering
%     \includegraphics[width = 0.8 \paperwidth]{pics/results_plots_obstacle_avoidance.png}
%     \caption{}
%     \label{fig:results_collision_avoidance}
% \end{subfigure}
% }
% \caption{(a) A static obstacle (black object) is detected in the red vehicle lane, the yellow and blue vehicles make gap for the red to merge into their lane. Vehicles are travelling in 2D configuration are flexible and are able to reshape in case of presence of obstacle in one lane. (b) The plots correspond to the snapshots and represents the vehicles' states and actions during the simulation. As seen, the steady state is achieved and 1D configuration is formed.}
% \end{figure*}

\subsection{Comparison with Behavior-Based Approach}
\begin{figure}
\vfill\makebox[\columnwidth][c]{
\begin{subfigure}[]{1\columnwidth}
    \centering
    \includegraphics[width = \columnwidth]{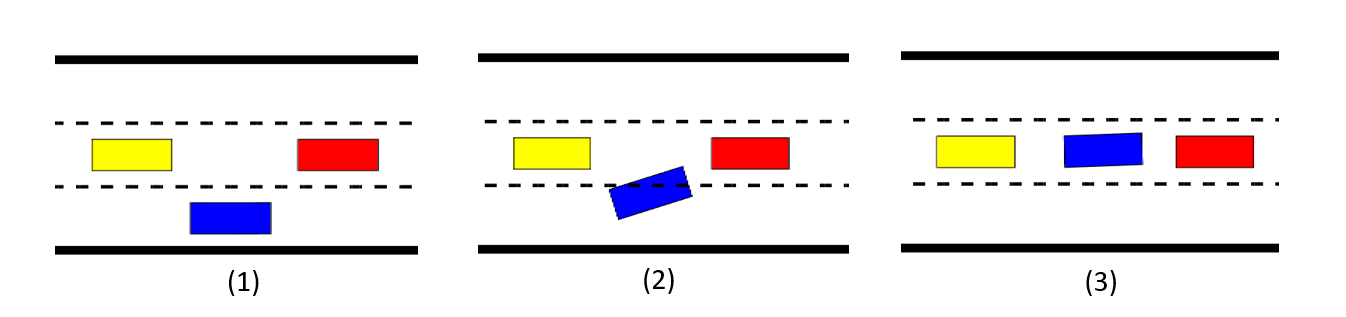}
    \caption{}
    \label{fig:results_sim_motion_primitives}
\end{subfigure}
}
\makebox[\columnwidth][c]{
\begin{subfigure}[]{1.2\columnwidth}
    \centering
    \includegraphics[width= \columnwidth]{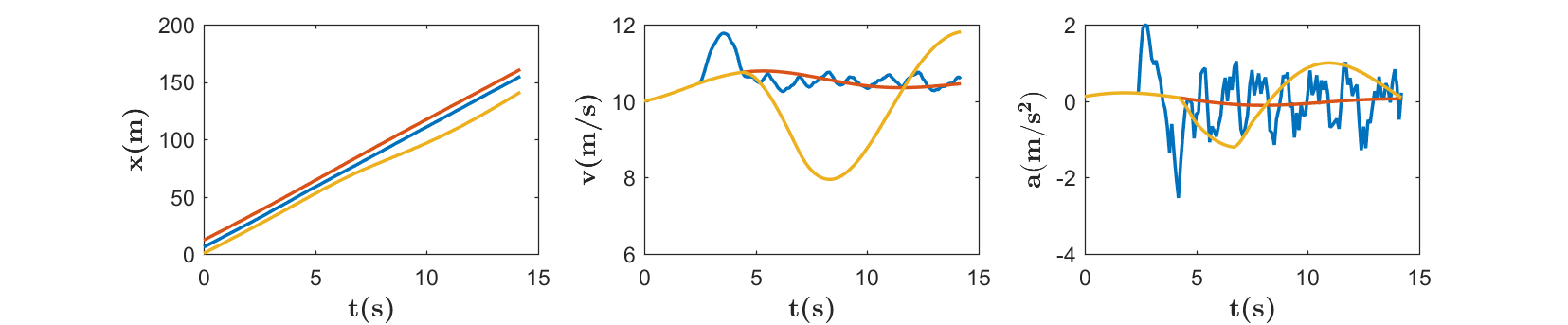}
    \caption{}
    \label{fig:plots_motion_primitves_compare}
\end{subfigure}
}
\hfill\makebox[\columnwidth][c]{
\begin{subfigure}[]{1.2\columnwidth}
   \centering
    \includegraphics[width=\columnwidth]{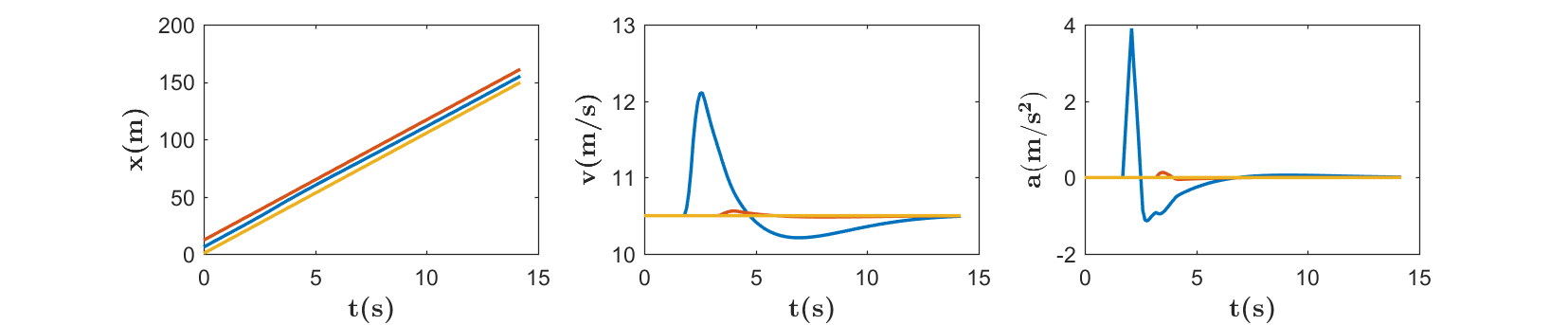}
    \caption{}
    \label{fig:plots_centralized_compare} 
\end{subfigure}
}
\caption{(a) Formation using sequence of motion primitives is demonstrated. At step (1), the cars are moving to the right in two-dimensional formation and the yellow car starts slowing down to make enough gap for the blue car to merge, while the red car doesn't change its speed. At step (2), the blue car changes lane to merge the platoon. At step (3) blue follows the red car and yellow follows the blue car and 1D platoon is formed. (b) Planning using sequence of motion primitives (c) Optimization-based planning.}
\end{figure}

To compare the proposed approach with the behavior-based approach discussed in \secref{secSimpleExample}, a simple example scenario is considered. Two vehicles, which are moving together in the same lane, make enough gap for the third vehicle to allow it to merge into their lane. This simple scenario is chosen to be able to determine the sequence of motion primitives for each vehicle intuitively without any mathematical analysis. However, sequence of motion primitives should be obtained using mathematical analysis such as MIP for more complicated scenarios. The simulation results for behavior-based approach is shown in \figref{fig:results_sim_motion_primitives}. The sequence of motion primitives for this simulation are:
\begin{enumerate}
    \item Red car follows a constant desired velocity (CC).  
    \item Yellow car slows down.
    \item Blue car performs lane-change.
    \item Blue car follows the red car (ACC).
    \item Yellow car follows the blue car (ACC).
\end{enumerate}
As seen in \figref{fig:results_sim_motion_primitives}, at step (1), the cars are moving to the right in two-dimensional platoon and the yellow car slows down to make a proper gap to allow the blue car to merge into the lane, while the red car is moving with constant speed. Step (2) shows the lane change of the blue car. Step (3) illustrates the reconfiguration of one-dimensional platoon. In this simulation, the collision avoidance constraints among the cars are not imposed, so the blue car changes its lane only after a large enough gap is created between the red and yellow cars. Even for this simple scenario obtaining maneuvers with larger velocity and closer inter-vehicle distance was impossible after running extensive simulations. The same scenario is replicated with optimization-based planning. The same initial conditions and parameters are used for both methods. The initial longitudinal coordinates for the three vehicles are $[x^1(0),x^2(0),x^3(0)]=[6,12,0.5]$ and the initial lateral coordinates are $[y^1(0),y^2(0),y^3(0)]=[1.85,5.55,5.55]$. $d_\text{min}$ is chosen as $0.2$m, the horizon $N$ is 8, the simulation sampling time $\Delta t$ is $0.1$s, simulation time $T$ is $150$, $\rho$ is $0.25$ and $v_{\text{max}}$ is $10.5$m/s. The resulting maneuvers obtained by motion primitive and optimization-based approaches are presented in \figref{fig:plots_motion_primitves_compare} and \figref{fig:plots_centralized_compare}, respectively. As seen in \figref{fig:plots_motion_primitves_compare}, the $x$ plot, the yellow car longitudinal position is far behind the other two. Also in $v$ plot, the yellow car reduces its speed dramatically and the blue car is changing its speed. However in \figref{fig:plots_centralized_compare}, that shows the obtained trajectories using optimization-based approach the cars maintain a tight inter vehicle distance as seen in the $x$ plot and the velocities and accelerations are changing smoothly. 

% \begin{figure*}
% \makebox[\linewidth][c]{20
% \begin{subfigure}[a]{2\paperwidth}
%     \centering
%     \includegraphics[width = 0.5\paperwidth]{pics/motion_primitives.png}
%     \caption{}
%     \label{fig:results_sim_motion_primitives}
% \end{subfigure}
% }
% \makebox[\linewidth][c]{20
% \begin{subfigure}[b]{1\paperwidth}
%     \centering
%     \includegraphics[width= 0.8\paperwidth]{pics/plots_motion_primitive_compare.png}
%     \caption{}
%     \label{fig:plots_motion_primitves_compare}
% \end{subfigure}
% }
% \makebox[\linewidth][c]{20
% \begin{subfigure}[c]{1\paperwidth}
%   \centering
%     \includegraphics[width=0.8\paperwidth]{pics/plots_centralized_compare.png}
%     \caption{}
%     \label{fig:plots_centralized_compare} 
% \end{subfigure}
% }
% \caption{(a) Formation using sequence of motion primitives is demonstrated. At step (1), the cars are moving to the right in two-dimensional formation and the yellow car starts slowing down to make enough gap for the blue car to merge, while the red car doesn't change its speed. At step (2), the blue car changes lane to merge the platoon. At step (3) blue follows the red car and yellow follows the blue car and 1D platoon is formed. (b) Planning using sequence of motion primitives (c) Optimization-based planning.}
% \end{figure*}

\begin{figure}
\centering
\makebox[\linewidth][c]{\includegraphics[width=1.2\columnwidth]{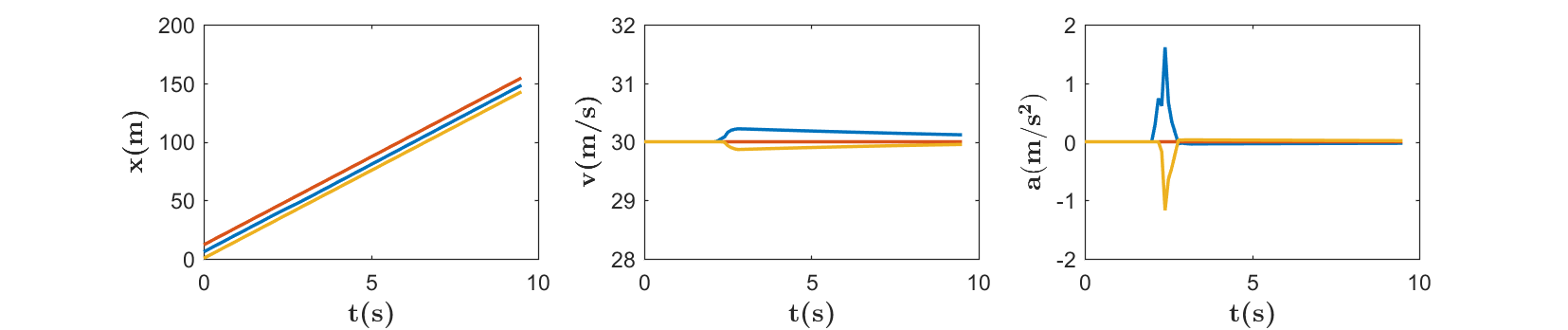}}
\caption{The trajectories obtained by optimization-based approach with highway speed and tight inter-vehicle distance are shown.}
\label{fig:centralized_v30}
\end{figure}

In addition, for this example, despite extensive tuning efforts, it was not possible to obtain trajectories at highway speed and tight inter-vehicle distance, using motion primitive approach. The reason is that this approach requires proper tuning of many parameters and switches as discussed earlier. However, the optimization-based approach yields trajectories with highway speed $30$m/s and tight inter-vehicle distance $0.2$m. The results are shown in \figref{fig:centralized_v30}. In summary, the motion primitive approach does not provably enforce the collision avoidance constraints. Furthermore, to design tight mini platoons at highway speed, the proposed optimization-based approach is simplified compared to motion primitives approach in which extensive tuning is required for all the switches and all the possible parameters. 

% \begin{figure*}
% \centering
% \makebox[0pt]{\includegraphics[width=0.8\paperwidth]{pics/plots_centralized_three_vehicle_comparison_v30.png}}
% \caption{The trajectories obtained by optimization-based approach with highway speed and tight inter-vehicle distance are shown.}
% \label{fig:centralized_v30}
% \end{figure*}

\subsection{Path Following}
\begin{figure}
\centering
\makebox[\linewidth][c]{\includegraphics[width=0.9\columnwidth]{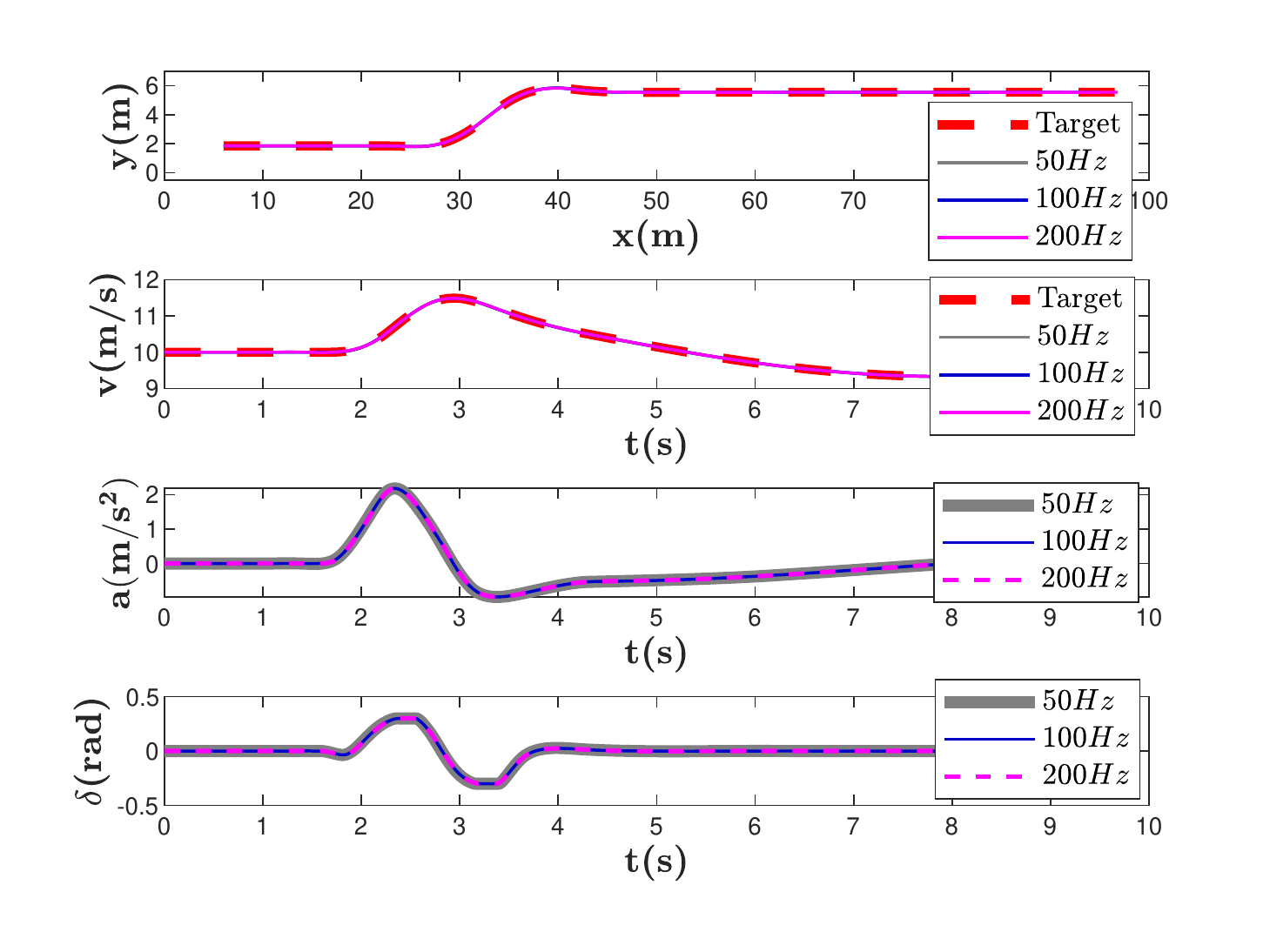}}
\caption{The closed-loop simulation of MPC path-following controller is shown for various sampling times.}
\label{fig:result_path_following}
\end{figure}

In this section, the MPC path-follower controller \eqref{eq:path_follower} is simulated in closed loop with dynamic model \eqref{eq:kinematic_bicycle_model}. The results are shown for a lane-change maneuver selected from the look-up table. The lane-change maneuver is planned by the motion-planner and path-follower follows the pre-computed motion. The results are reported in \figref{fig:result_path_following} for various sampling rates including $50 Hz$, $100 Hz$ and $200 Hz$. The top plot shows the path in $xy$ plane. The target trajectory (obtained by motion-planner) is shown with red dashed line. The gray, blue and pink plots are the results of path-follower controller with different sampling rates. The second plot shows the velocity tracking, in which the red dashed line is the target trajectory obtained by motion planner. The third and forth plots are acceleration and steering angle, respectively, which are obtained by the MPC path-follower. The results show that path tracking and velocity tracking performance are not affected by changing the sampling rate. However, \ref{tab:computation_cost_sampling} compares the average and maximum of the computation time for different sampling rates. As seen, the average of computation time is reduced for lower sampling rate.

\begin{table}[]
\caption{Computation time in seconds for various sampling rates}\label{tab:computation_cost_sampling}
\centering
\begin{tabular}{|c|c|c|}
\hline
\textbf{Sampling Rate $(Hz)$}& \textbf{Average $(s)$} & \textbf{Max. $(s)$} \\
\hline
50 & 0.30 & 0.9 \\
100 & 0.42 & 1.91 \\
200 & 1.04 & 5.24 \\
\hline
\end{tabular}
\end{table}

In addition, six different simulations have been run (with sampling rate of 50 Hz) and the average and maximum of computation time of the controller is reported (in seconds) at Table \ref{tab:computation_cost}. These results are reported by running the simulation on a Surface Book laptop with Intel(R) Core(TM) i7-6600U CPU @2.81 GHz and 16.0GB RAM in MATLAB. The total average of the computation time of the MPC controller is $0.32 (s)$ and the maximum is $0.61 (s)$. Note that these values can be reduced dramatically (an order of magnitude) if the controller's dynamic model \eqref{dynamic_path_following} is linearized around the given target trajectory. The linearized version of \eqref{eq:path_follower} can be solved in real-time.

\begin{table}[]
\caption{Computation time in seconds for sampling rate of $50 Hz$}\label{tab:computation_cost}
\centering
\begin{tabular}{|c|c|c|}
\hline
\textbf{Run $\#$}& \textbf{Average $(s)$} & \textbf{Max. $(s)$} \\
\hline
1 & 0.31 & 0.71 \\
2 & 0.21 & 0.55 \\
3 & 0.45 & 0.68 \\
4 & 0.28 & 0.52 \\
5 & 0.33 & 0.61 \\
6 & 0.35 & 0.59\\
\hline
\textbf{Total Avg.} & 0.32 & 0.61\\
\hline
\end{tabular}
\end{table}

\section{Conclusion}
\label{secConclusion}
An architecture for autonomous navigation of multi-lane platoons on public roads is proposed. The architecture is composed of an offline motion-planning system and an online hierarchical control system, which consists of TOS, decision-maker and path-follower. The motion-planner avoids collisions among the vehicles within the platoon, but does not consider the collisions with surrounding vehicles outside the platoon. However, decision-maker checks the possible collisions between the planned reconfiguration maneuver and the future planned trajectories of the surrounding vehicles shared via V2V communication. Once a feasible reconfiguration maneuver is selected by the decision-maker, it will be executed by the path-follower controller in real time. The simulation results demonstrate that a platoon of vehicles can form geometrically flexible and reconfigurable shapes in tight environment while moving at highway speed. It is shown that in the case of sudden change in the environment, like appearing an obstacle or slow traffic in one lane, the multi-lane platoon of vehicles can perform collaborative maneuvers and change their configuration to merge into faster lanes. The proposed approach is compared with behavior-based planning, in which the formation and reconfiguration is achieved by a sequence of motion primitives. The results show that to design tight maneuvers for mini-platoons at highway speed the proposed optimization-based method is simplified compared to the motion primitive approach, which requires extensive tuning for the switches and parameters. The future work will be robustification of the planning scheme by handling the uncertainty caused by model mismatch, sensor measurements and communication delays and using closed-loop policies instead of open-loop ones.

\section{Acknowledgement}
\label{secAcknowledgement}
The information, data, or work presented herein was funded in part by the Advanced Research Projects Agency-Energy (ARPA-E), U.S. Department of Energy, under Award Number DE-AR0000791. The views and opinions of authors expressed herein do not necessarily state or reflect those of the United States Government or any agency thereof.

\bibliographystyle{IEEEtran}
\bibliography{main}

\end{document}